%% file: main.tex
\documentclass[letterpaper,journal]{IEEEtran}

\usepackage{cite}
\usepackage{amsmath,amssymb,amsfonts}
\usepackage{graphicx}
\usepackage{booktabs}
\usepackage{array}
\usepackage{multirow}
\usepackage{makecell}
\usepackage[table]{xcolor}
\usepackage{tikz}
\usepackage{url}

\usetikzlibrary{arrows.meta,positioning}
\begin{document}

\title{SPLG-Mamba: Structure-Preserving Local-Global Mamba Network for Salient Object Detection in Optical Remote Sensing Images}

\author{Yi~Xu, Ruichao~Hou,~\IEEEmembership{Member,~IEEE,} Tongwei~Ren,~\IEEEmembership{Member,~IEEE,} and Gangshan~Wu,~\IEEEmembership{Member,~IEEE}%
\thanks{This work was supported by the National Natural Science Foundation of China (No. 92582103), the Fundamental and Interdisciplinary Disciplines Breakthrough Plan of the Ministry of Education of China (No. JYB2025XDXM118), the ``111 Center'' (No. B26023), and the Collaborative Innovation Center of Novel Software Technology and Industrialization. \textit{(Corresponding author: Tongwei Ren.)}}%
\thanks{Yi Xu, Tongwei Ren, and Gangshan Wu are with the State Key Laboratory for Novel Software Technology, Nanjing University, Nanjing, 210008, Jiangsu, China (e-mail: yxu1025@smail.nju.edu.cn; rentw@nju.edu.cn; gswu@nju.edu.cn).}%
\thanks{Ruichao Hou is with the School of Elite Biomedical Engineers and the Institute for Interdisciplinary Intelligent Pharmacy, China Pharmaceutical University, Nanjing 211198, China (e-mail: rchou@cpu.edu.cn).}%
}


\maketitle

\begin{abstract}
Salient object detection in optical remote sensing images (ORSI-SOD) requires dense predictions that preserve object completeness and structural continuity under complex backgrounds, scale variation, and irregular object shapes. Existing methods often localize salient regions, but their predictions may still suffer from structural degradation, including fragmented, incomplete, or locally missing foreground responses. This degradation is closely related to hierarchical feature propagation, where shallow details can introduce texture-induced background responses, deep semantics may over-smooth weak structures, and uncontrolled cross-scale fusion can disturb coherent regions. To address this issue, we propose a novel Structure-Preserving Local-Global Mamba Network, SPLG-Mamba, for ORSI-SOD. Specifically, SPLG-Mamba integrates Smooth-Detail Recalibration (SDR), hierarchy-aware Local-Global Mamba, and Gated Cross-Scale Fusion (GCSF). SDR recalibrates smoothed responses and detail residuals before state-space modeling, Local-Global Mamba assigns local modeling to shallow feature levels and global modeling to deep feature levels, and GCSF controls cross-scale detail injection during decoding. Experiments on ORSSD, EORSSD, and ORSI-4199 demonstrate state-of-the-art results and improved structural completeness and continuity. The code is available at \url{https://github.com/yxu9910/SPLG-Mamba}.
\end{abstract}


\section{Introduction}

\IEEEPARstart{S}{alient} object detection (SOD) aims to segment the most visually prominent objects or regions in an image. In optical remote sensing images, SOD provides useful saliency cues for geographic interpretation, target localization, scene understanding, and human-in-the-loop analysis. Nevertheless, extending SOD methods developed for natural scene images to ORSI-SOD is nontrivial. Optical remote sensing images are captured from overhead viewpoints, cover large fields of view, and contain salient targets with diverse scales, orientations, shapes, and foreground-background contrasts~\cite{li2019orssd,zhang2021eorssd,liu2023orsi4199}. These characteristics can fragment salient regions, suppress weak foreground parts, or confuse targets with structured backgrounds, making structural preservation a central challenge in ORSI-SOD. For ORSI-SOD, structural preservation concerns the completeness of predicted salient regions and the continuity of narrow or connected object parts.

Existing ORSI-SOD methods have improved saliency prediction from several complementary directions. Multiscale and feature-fusion methods use shallow details and deep semantics to handle large-scale variation~\cite{li2019orssd,li2022mccnet}. Region-boundary methods strengthen foreground regions and object contours to reduce incomplete or blurred predictions~\cite{tu2022mjrbm,feng2023bscgnet,lee2024lshnet}. Context-oriented methods, including Transformer-based, local-global, and Mamba-based designs, enhance long-range dependency modeling and global scene understanding~\cite{sun2025lgipnet,yang2025thmnet,xing2025lightemnet}. These advances have substantially strengthened ORSI-SOD, especially in foreground localization and scene-level discrimination.

\begin{figure}[!t]
\centering
\includegraphics[width=\columnwidth]{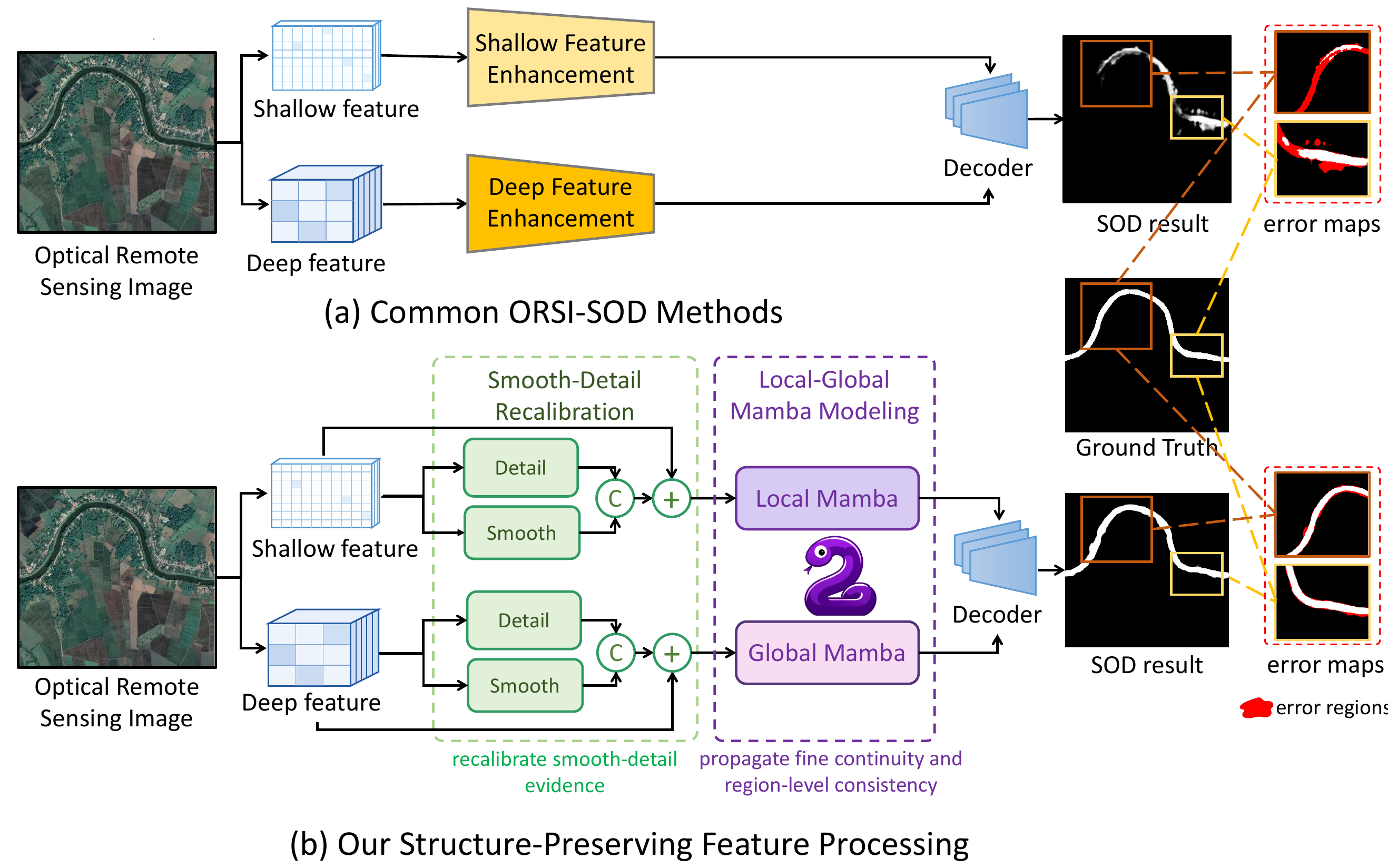}
\caption{Feature-processing illustration for ORSI-SOD. The upper row abstracts a representative encoder-decoder pipeline for ORSI-SOD. The lower row illustrates the proposed smooth-detail recalibration before Local-Global Mamba. White, black, and red indicate correctly predicted foreground, correctly predicted background, and prediction errors, respectively.}

\label{fig:intro-motivation}
\end{figure}

However, structural degradation can still appear during hierarchical feature propagation before the final prediction. Many existing ORSI-SOD methods enhance shallow and deep features separately and integrate them during decoding~\cite{feng2023bscgnet,zeng2023aesinet,lee2024lshnet,sun2025lgipnet}. This distinction reflects two feature properties of optical remote sensing images: shallow features preserve local structures but are sensitive to background textures from roofs, roads, and water surfaces, whereas deep features provide object-level semantics but may suppress weak or narrow foreground parts. Yet distinguishing feature levels alone does not determine how region-level cues and local structure cues should be coordinated during hierarchical feature propagation. Without sufficient coordination, shallow details may introduce texture-induced background responses, whereas over-smoothed deep semantics may weaken fine foreground structures. As shown in Fig.~\ref{fig:intro-motivation}, SDR recalibrates the smoothed response and detail residual of each feature before Local-Global Mamba modeling. Local Mamba restricts shallow-feature propagation to local windows, whereas Global Mamba models complete deep feature maps to maintain object-level consistency.

To address this limitation, we propose SPLG-Mamba, a Structure-Preserving Local-Global Mamba Network for ORSI-SOD. Mamba provides efficient long-range feature propagation, but its performance in ORSI-SOD is affected by the feature evidence entering state-space modeling and the spatial scope of propagation. SPLG-Mamba addresses these two factors through SDR and hierarchy-aware Local-Global Mamba.

SPLG-Mamba consists of three coordinated components for structure-preserving feature propagation. First, SDR separates encoder features into smoothed responses and detail residuals before Mamba modeling. The smoothed response provides region-level cues, while the detail residuals preserve local structure cues for state-space modeling. Second, Local-Global Mamba assigns Local Mamba blocks to shallow high-resolution features and Global Mamba blocks to deep semantic features, so that each feature level uses a suitable modeling scope. Third, GCSF uses a gating mechanism to control lateral detail injection into the top-down semantic stream, preserving detail structures that are consistent with high-level saliency semantics during decoding.

The main contributions are summarized as follows:
\begin{itemize}
    \item We propose SPLG-Mamba, a structure-preserving ORSI-SOD framework. To the best of our knowledge, it is the first Mamba-based ORSI-SOD framework to organize state-space modeling across the feature hierarchy, assigning local modeling to shallow detail-sensitive features and global modeling to deep semantic features.
    \item We design SDR and GCSF to support structure-preserving propagation before and after Mamba modeling. SDR recalibrates smoothed responses and detail residuals before state-space modeling, while GCSF controls cross-scale detail injection during decoding to preserve semantically consistent detail structures.
    \item Extensive experiments on ORSSD, EORSSD, and ORSI-4199 demonstrate that SPLG-Mamba achieves state-of-the-art performance and consistently improves structural completeness and continuity in challenging remote-sensing scenes.
\end{itemize}

\section{Related Work}

\subsection{Boundary and Structure Guidance}

Boundary and structure guidance is widely used in SOD to recover complete salient regions rather than only highly discriminative object parts~\cite{zhuge2023integrity}. This requirement is more difficult in ORSI-SOD, where weak contrast, irregular contours, and complex backgrounds can make object extents ambiguous. MJRBM jointly models salient regions and object boundaries to reduce incomplete foreground responses~\cite{tu2022mjrbm}. AESINet introduces edge-aware semantic interaction, while LSHNet updates hierarchical features with structure priors~\cite{zeng2023aesinet,lee2024lshnet}. RAGRNet and MRBINet further strengthen region-boundary interaction through recurrent graph reasoning and multistrategy fusion, respectively~\cite{zhao2024ragrnet,jia2025mrbinet}. DAFNet, BSCGNet, and ERNet introduce dense attention, boundary-semantic collaboration, or edge-guided refinement to sharpen object transitions~\cite{zhang2021eorssd,feng2023bscgnet,wang2025ernet}. More recently, integrity-detail ensemble learning considers the balance between complete salient regions and fine details~\cite{liu2024ensemble}.

These methods demonstrate that boundary and structure guidance is useful for recovering object contours and foreground extents, establishing structural completeness as an important objective for ORSI-SOD. Beyond explicit contour recovery, structural degradation can also arise during hierarchical feature propagation before the final saliency map is decoded. In this process, shallow texture responses, deep semantic features, and lateral feature fusion may interact before final decoding. Structural errors can therefore remain even when the object is roughly localized, especially for thin, incomplete, or internally complex salient regions.

\subsection{Multi-Scale Feature Fusion and Frequency-Aware Modeling}

Multi-scale feature fusion addresses scale variation by combining coarse semantic localization with fine spatial recovery. LVNet uses a two-stream pyramid and nested decoding to handle salient objects at different scales~\cite{li2019orssd}. ACCoNet coordinates adjacent contexts, MCCNet complements multiple feature contents, and MFENet enhances multiscale features for stronger encoder-decoder interaction~\cite{li2022adjacent,li2022mccnet,wang2022mfenet}. PFIFNet further interleaves features progressively so that shallow and deep features can be refined across decoding stages~\cite{han2024pfifnet}. These designs show that effective ORSI-SOD requires the joint use of high-resolution details and low-resolution semantics, especially for scenes containing small objects, large objects, elongated structures, and repeated background patterns.

Frequency-aware modeling provides another way to separate region-level layout from local structural variations. Frequency-tuned saliency separates slowly varying layout responses from edge- and texture-sensitive components~\cite{achanta2009frequency}. UDCNet uses frequency-spatial information for domain cognition, and FreMaNet introduces frequency-domain self-attention for efficient intralevel relation modeling~\cite{sun2024udcnet,li2026frema}. IPDiff combines reconstructed hierarchical priors with spatial-spectral guidance in a conditional diffusion framework~\cite{li2026ipdiff}. These frequency-aware responses are useful for representing fine structures, but they are also ambiguous in optical remote sensing images: roads, roofs, and shorelines can be as strong as true object details.

Multi-scale feature fusion and frequency-aware modeling provide useful feature representations for objects with diverse sizes and fine structures. The remaining difficulty lies in how these multiscale and frequency-aware features are propagated through the hierarchy. Local detail responses may introduce structured background noise, while deep semantics may over-smooth weak foreground parts. Structural preservation therefore requires smooth-detail recalibration and controlled cross-scale propagation, so that useful local structures can support foreground continuity without disturbing stable semantic regions.

\begin{figure*}[t]
\centering
\includegraphics[width=\textwidth]{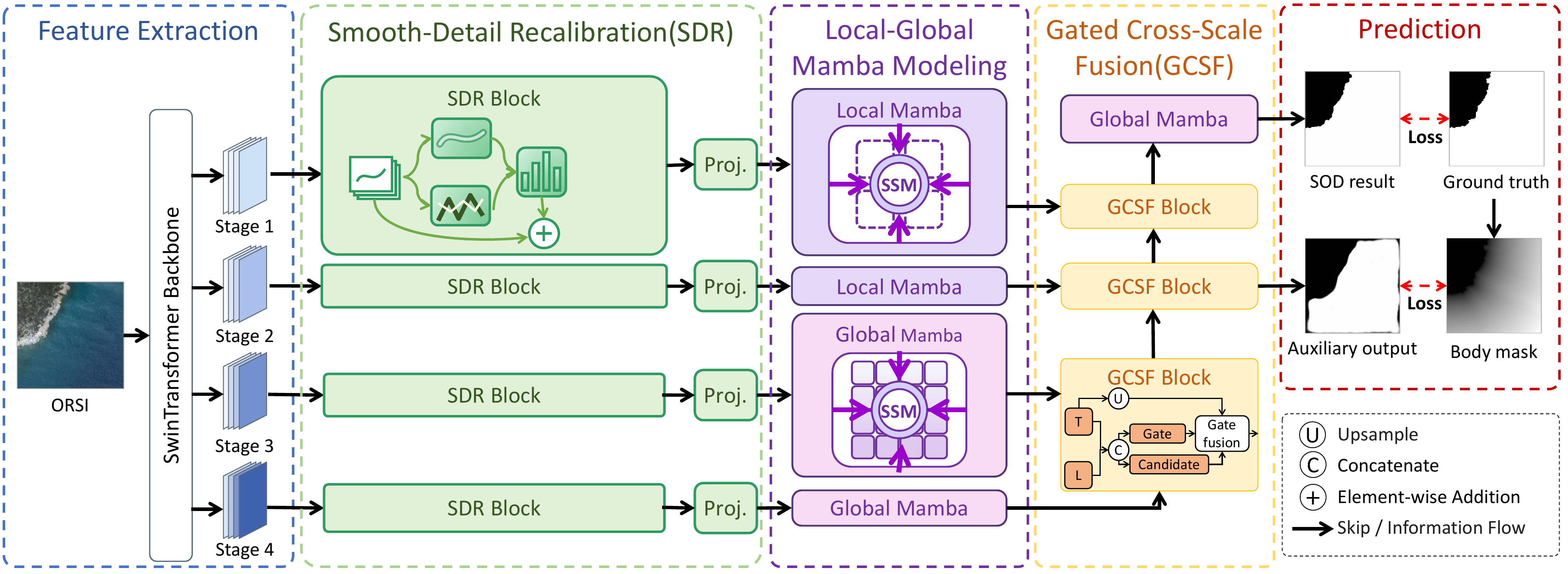}
\caption{The framework of SPLG-Mamba. SPLG-Mamba consists of a SwinV2 encoder, SDR, Local-Global Mamba, GCSF, and prediction heads.}
\label{fig:pipeline}
\end{figure*}

\subsection{Local-Global Context Modeling}

Context modeling helps distinguish salient objects from confusing backgrounds by using semantic interaction and long-range dependencies. GeleNet introduces Transformer-driven saliency prediction, and ADSTNet improves feature interaction through adaptive sparse tokenization~\cite{li2023gelenet,10439066}. HFCNet models heterogeneous feature collaboration for more robust scene understanding~\cite{liu2024hfcnet}. Several recent methods further model local-global context. LGIPNet explicitly models local-global information perception, TLCKDNet uses a large-kernel decoder to enlarge the effective receptive field, and DPUFormer introduces a dual-perspective Transformer for object segmentation~\cite{sun2025lgipnet,DONG2024103917,sun2025dpuformer}. CMNFNet combines CNN-based local patterns with graph-based local and global receptive fields through nested heterogeneous-feature fusion~\cite{xu2025cmnfnet}.

These methods are effective for localization and scene-level discrimination, but local and global interactions are often used as general context modules or decoder components. They usually do not specify which feature levels should favor local interaction and which levels should favor global interaction. Their fusion stages also seldom specify how lateral details enter the top-down semantic stream. As a result, global context may suppress small structural parts, whereas shallow details may disturb stable semantic regions during fusion.

\subsection{Mamba-Based Modeling}

Mamba is a selective state-space model that captures long-range dependencies with linear complexity~\cite{gu2024mamba}. Vision Mamba, VMamba, and LocalMamba extend state-space modeling to visual representation learning by combining efficient sequence propagation with spatial modeling capability~\cite{visionmamba,liu2024vmamba,localmamba2024}. This property is attractive for dense prediction because object continuity may need to be modeled across a large spatial extent without the quadratic cost of full self-attention.

Recent ORSI-SOD methods have begun to use this modeling ability. LightEMNet uses a Mamba-based cross-scale edge-semantic interaction module for weakly supervised saliency detection~\cite{xing2025lightemnet}. THMNet uses grouped Mamba units for multidirectional global modeling and three Mamba heads to refine foreground, background, and global features~\cite{yang2025thmnet}. MambaCVC adopts a VMamba-style backbone and compensates correlation variance between global and local contexts for efficient ORSI-SOD~\cite{di2026ldcvc}. Diffusion-distillation methods also incorporate Vision Mamba-based spatial modeling into teacher-student generation pipelines~\cite{gao2026diffdistill}. These studies demonstrate the potential of state-space modeling for remote-sensing saliency. Their Mamba components mainly support edge-semantic interaction, global topology modeling, or backbone feature extraction. The assignment of local and global Mamba modeling to shallow and deep features, respectively, has received less attention. We therefore recalibrate smoothed responses and detail residuals before state-space propagation, apply local modeling to shallow features and global modeling to deep features, and control cross-scale detail injection during decoding to preserve local structures and coherent foreground regions.

\section{Method}

\subsection{Overview}

As shown in Fig.~\ref{fig:pipeline}, SPLG-Mamba is organized around structure-preserving hierarchical feature propagation. Given an input image, a SwinV2 encoder~\cite{liu2022swinv2} extracts a four-level feature hierarchy $\{F_i\}_{i=1}^{4}$, ranging from shallow high-resolution features to deep low-resolution features. SDR then recalibrates each encoder feature to produce $\{\widehat{F}_i\}_{i=1}^{4}$ before Mamba modeling. The recalibrated features are projected to a common channel dimension and processed by Local-Global Mamba to obtain $\{Z_i\}_{i=1}^{4}$, where shallow levels are modeled locally and deep levels globally. Starting from $Z_4$, GCSF progressively controls lateral detail injection from $Z_3$, $Z_2$, and $Z_1$ into the top-down stream before the final SOD result is predicted. During training, an auxiliary body output is supervised by a body mask generated from the ground-truth saliency annotation.

This organization places structure-preserving operations at successive stages of hierarchical feature propagation. Specifically, SPLG-Mamba recalibrates encoder features before state-space modeling, uses Local-Global Mamba to assign local modeling to shallow features and global modeling to deep features, and controls cross-scale detail injection during decoding; the following subsections describe these operations and the training objective in this order.

\subsection{Smooth-Detail Recalibration}

SDR recalibrates encoder features by separating two complementary components: smoothed responses and detail residuals. Smoothed responses provide region-level cues for coherent object areas, whereas detail residuals preserve boundaries, thin parts, and internal structures. In optical remote sensing images, similar local patterns also appear in roofs, roads, and water surfaces, making detail residuals sensitive to background textures. If these residuals are propagated without recalibration, background textures can be amplified; if they are suppressed too early, narrow or incomplete salient regions can be lost.

\begin{figure}[t]
\centering
\includegraphics[width=\columnwidth]{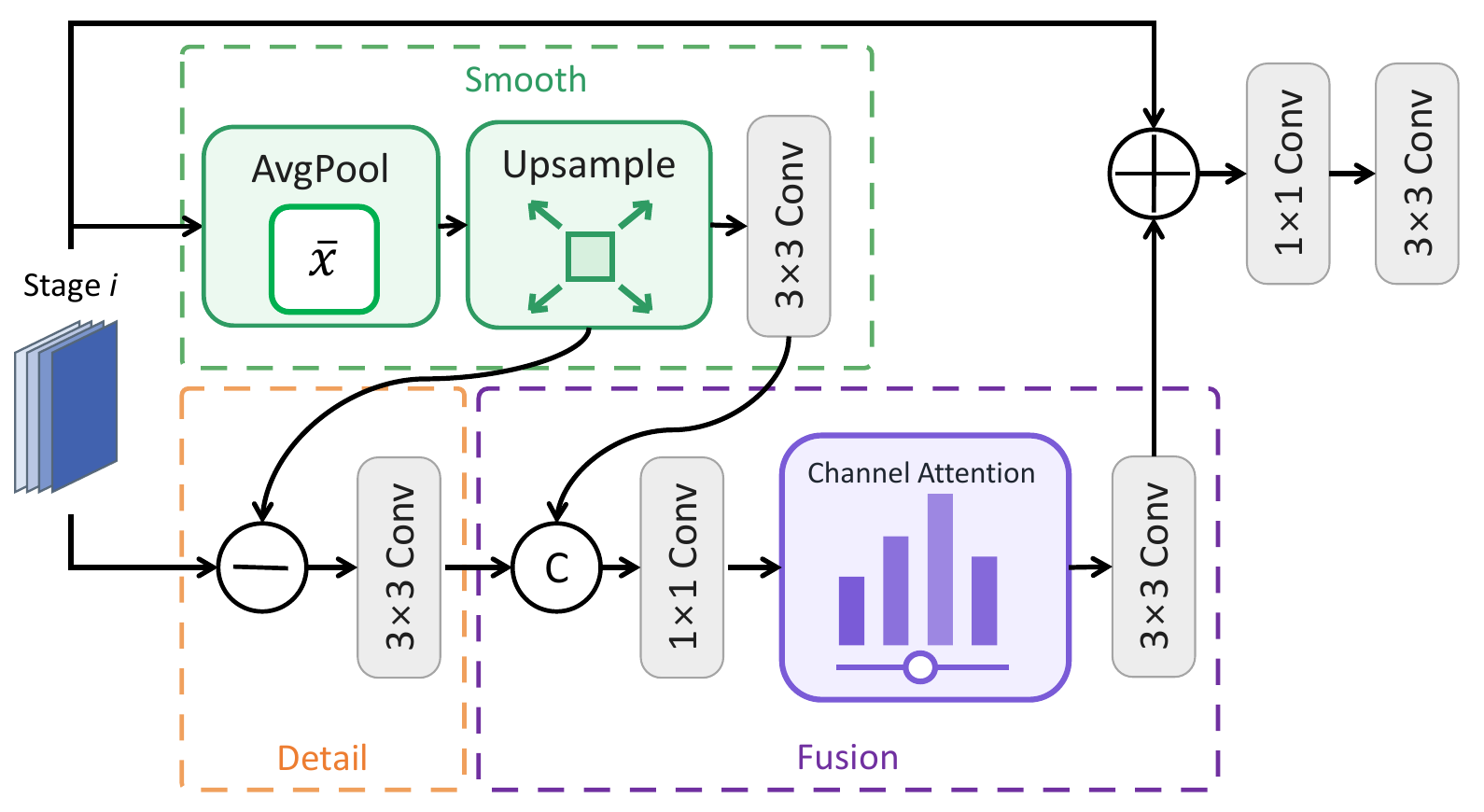}
\caption{Details of the SDR block.}
\label{fig:sdr}
\end{figure}

Figure~\ref{fig:sdr} shows the details of the SDR block. The block uses a feature-splitting operation~\cite{lu2022esrt} to form a smoothed response and a detail residual, compares them through convolutional filtering, and uses residual channel attention~\cite{zhang2018rcan} to produce a correction. In SPLG-Mamba, this recalibration is placed before Mamba modeling so that subsequent state-space modeling receives balanced region-level cues and local structure cues.

Concretely, for the $i$-th encoder feature $F_i$, the SDR block first obtains a smoothed response $A_i$ by applying $2\times2$ average pooling and then restoring the original resolution by bilinear interpolation. The detail residual $R_i$ is obtained by subtracting $A_i$ from $F_i$. At each encoder level, $R_i$ represents spatial variations relative to the corresponding smoothed feature $A_i$. A learned correction path $\Delta_i(\cdot)$ compares $R_i$ and $A_i$ through convolution and channel attention, and the correction is added back to $F_i$ to produce the recalibrated feature $\widehat{F}_i$. This residual form keeps $\widehat{F}_i$ anchored to the encoder representation, while allowing the network to suppress local residuals that lack region-level cues. By applying SDR blocks across the encoder hierarchy, SDR preserves object details for later modeling while reducing unsupported local activations before state-space modeling.

\subsection{Local-Global Mamba}

The recalibrated features then require hierarchy-specific state-space modeling. Shallow high-resolution features contain fine structures and repeated background patterns, so full-map interaction may connect unrelated local responses under cluttered backgrounds. Deep low-resolution features are more semantic and spatially compact, making global modeling more suitable for object-level organization.

\begin{figure}[t]
\centering
\includegraphics[width=0.9\columnwidth]{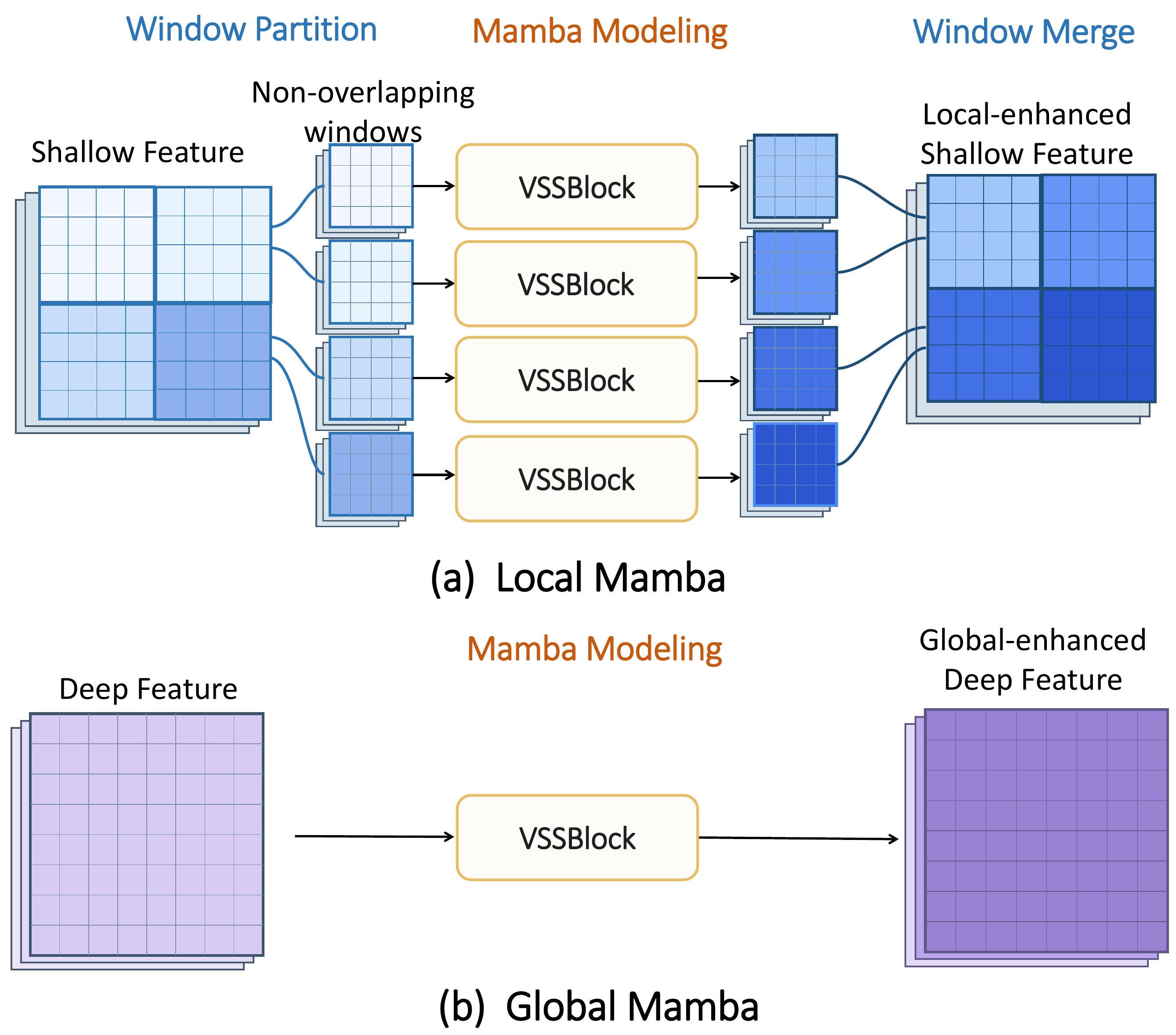}
\caption{Illustration of Local-Global Mamba.}
\label{fig:mamba}
\end{figure}

Local Mamba and Global Mamba use the same state-space operator with different spatial scopes. Both Local Mamba and Global Mamba adopt the Visual State-Space Block (VSSBlock) from VMamba~\cite{liu2024vmamba}, where 2D Selective Scan aggregates spatial context along multiple scanning routes. Specifically, the scan traverses row-wise and column-wise sequences in both forward and reverse directions. With VSSBlock as the shared state-space operator, shallow features are processed within local windows, whereas deep features are processed over the complete spatial map, as shown in Fig.~\ref{fig:mamba}.

Before Mamba modeling, each recalibrated feature is aligned to a common channel dimension using a $1\times1$ projection and a $3\times3$ convolution block. Local Mamba processes the first two aligned levels by partitioning each feature into non-overlapping local windows, applying VSSBlock independently inside each window, and restoring the windows to their original positions to obtain $\{Z_i\}_{i=1}^{2}$. Global Mamba processes the last two aligned levels by applying the same VSSBlock over the complete spatial map of each level to obtain $\{Z_i\}_{i=3}^{4}$. This assignment supports structural preservation by matching modeling scope to feature properties: local modeling keeps shallow details from spreading across cluttered regions, while global modeling gives deep semantic features enough spatial reach to maintain complete salient objects.

\subsection{Gated Cross-Scale Fusion}

GCSF controls lateral detail injection during decoding. Lateral features are important for recovering boundaries and small structures, but they may also contain texture responses from cluttered backgrounds. The top-down stream provides more stable semantic guidance but remains spatially coarse, whereas the lateral stream contains richer spatial details with lower semantic reliability. The semantic-spatial mismatch between the top-down and lateral streams requires selective detail injection. GCSF therefore performs gated cross-scale fusion, injecting lateral details according to their compatibility with the top-down stream.

At a decoding level, a GCSF block updates a top-down decoder feature $T$ with a same-level lateral Mamba feature $L$. It upsamples $T$ to the resolution of $L$, producing $\widetilde{T}=\mathrm{Up}(T)$. The two features are concatenated as $Q=\mathrm{Concat}(\widetilde{T},L)$. A fusion branch generates a candidate update $U$ from $Q$ through global channel reweighting, channel reduction, and spatial selection. The GCSF block also predicts a sigmoid gate from both $\widetilde{T}$ and $L$, and uses this gate to combine the candidate update with the top-down stream.

Specifically, the gate and fused output are computed as
\begin{align}
    \mathbf{g} &= \mathrm{Sigmoid}\big(\mathrm{BN}(\mathrm{Conv}_{1\times1}(Q))\big), \\
    Y &= \mathrm{Conv}_{3\times3}\big(\mathbf{g}\odot U+(1-\mathbf{g})\odot\widetilde{T}\big),
\end{align}
where $Y$ is the fused output, $U$ is the candidate update derived from cross-scale features, $\mathbf{g}\in[0,1]^{C\times H\times W}$ is the sigmoid gate, $\odot$ denotes element-wise multiplication, and $\mathrm{BN}$ denotes batch normalization. Larger gate values favor the candidate update, while smaller values preserve the upsampled semantic stream. Starting from $Z_4$, GCSF blocks progressively fuse the lateral features $Z_3$, $Z_2$, and $Z_1$, allowing the decoder to update fragmented or incomplete regions where lateral details are useful while preserving coherent regions from unreliable shallow updates. After the three GCSF blocks, a Global Mamba block further refines the fused feature before it is passed to the prediction heads.

\subsection{Loss Function}

Following common SOD supervision settings, we train the final saliency output and an auxiliary body output jointly. Let $S$ denote the final saliency prediction and $S_b$ denote the auxiliary body prediction generated from an intermediate decoder feature. The auxiliary branch is used only during training and removed during inference. Its body-mask target $G_b$ is generated from the saliency annotation $G_s$ by a normalized distance transform, so that pixels farther from object boundaries receive stronger foreground-interior supervision.

The training objective is defined as
\begin{equation}
    \mathcal{L}_{sup}
    =\lambda_s\mathcal{L}_{mask}(S,G_s)+\lambda_b\mathcal{L}_{mask}(S_b,G_b),
\end{equation}
where $\lambda_s$ and $\lambda_b$ balance the final saliency supervision and the auxiliary body-mask supervision, respectively. For each prediction-target pair, $\mathcal{L}_{mask}$ combines intersection-over-union (IoU) loss~\cite{rahman2016iou}, Dice loss~\cite{milletari2016vnet}, and structural similarity index measure (SSIM) loss~\cite{wang2004ssim}:
\begin{equation}
    \mathcal{L}_{mask}
    =\mathcal{L}_{IoU}+\mathcal{L}_{Dice}+\mathcal{L}_{SSIM}.
\end{equation}
The IoU and Dice terms supervise foreground coverage, while the SSIM term encourages local structural consistency.

\section{Experiments}

\input{tables/overall_comparison.tex}

\begin{figure*}[!t]
\centering
\includegraphics[width=0.96\textwidth]{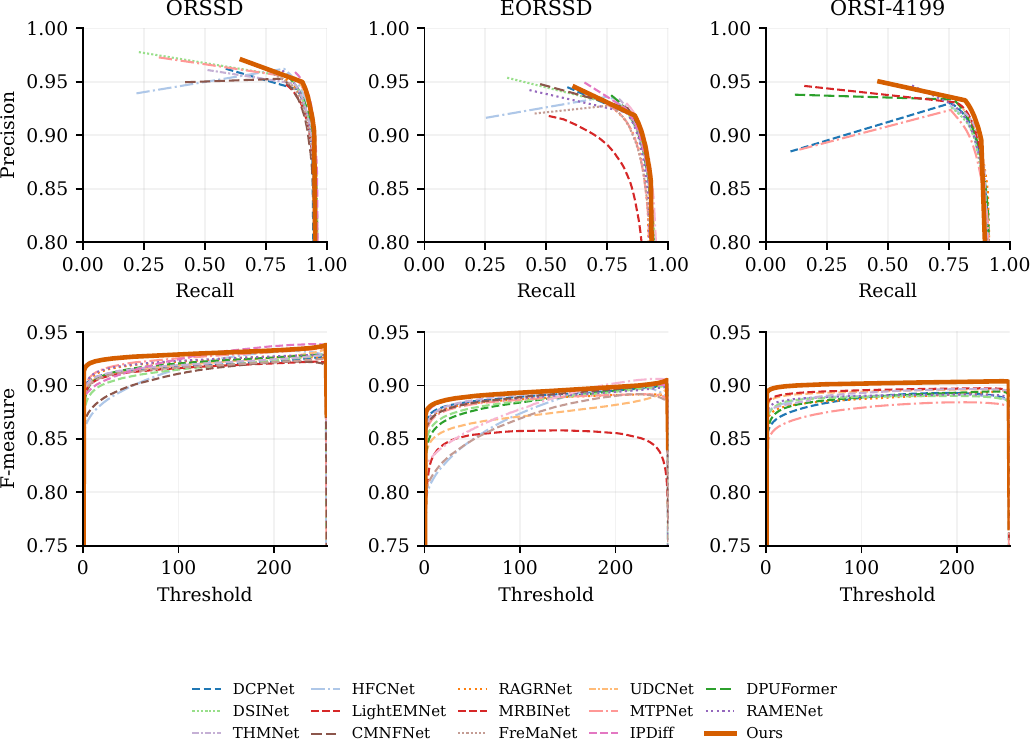}
\caption{Precision-recall and F-measure curves on ORSSD, EORSSD, and ORSI-4199.}
\label{fig:prf}
\end{figure*}

\input{tables/complexity.tex}

\subsection{Experimental Setup}

\textbf{Datasets.}
We evaluate SPLG-Mamba on three ORSI-SOD benchmark datasets: ORSSD, EORSSD, and ORSI-4199~\cite{li2019orssd,zhang2021eorssd,tu2022mjrbm}. ORSSD contains 600 training images and 200 test images, EORSSD contains 1,400 training images and 600 test images, and ORSI-4199 contains 2,000 training images and 2,199 test images. ORSI-4199 additionally provides annotations for nine attributes: big salient object (BSO), small salient object (SSO), off-center (OC), complex salient object (CSO), complex scene (CS), narrow salient object (NSO), multiple salient objects (MSO), low-contrast scene (LCS), and incomplete salient object (ISO).

\textbf{Metrics.}
Following common ORSI-SOD protocols, we report mean absolute error ($\mathcal{M}$), S-measure ($S_m$)~\cite{fan2017structuremeasure}, weighted F-measure ($F_\beta^w$)~\cite{margolin2014wfm}, maximum and mean F-measure ($F_\beta^{\max}$ and $\overline{F_\beta}$)~\cite{achanta2009frequency}, and maximum E-measure ($E_\xi^{\max}$)~\cite{fan2018emeasure}. Precision-recall and F-measure curves use 256 thresholds with $\beta^2=0.3$; $F_\beta^{\max}$ is the maximum over these thresholds, whereas $\overline{F_\beta}$ is their mean. The conventional metrics evaluate overall saliency prediction quality from complementary perspectives. Among them, $S_m$ measures object- and region-aware structural similarity, while $F_\beta^w$ accounts for spatial dependencies and error locations. To provide an additional measure of structural similarity across spatial scales, we report the multi-scale structural similarity index (MS-SSIM) on ORSI-4199~\cite{wang2003msssim}, computed as
\begin{equation}
\operatorname{MS\text{-}SSIM}(X,Y)
=\operatorname{SSIM}_{M}(X,Y)^{w_M}
\prod_{j=1}^{M-1}\operatorname{CS}_{j}(X,Y)^{w_j},
\end{equation}
where $X$ and $Y$ denote the prediction and ground truth, respectively, and $\operatorname{CS}_{j}$ is the contrast-structure comparison at scale $j$. We use $M=5$ and the standard weights $\{0.0448, 0.2856, 0.3001, 0.2363, 0.1333\}$. Lower $\mathcal{M}$ is better, whereas higher values indicate better performance for the other metrics.

\textbf{Implementation details.}
SPLG-Mamba is implemented in PyTorch~\cite{paszke2019pytorch}. During training, we apply random horizontal flipping, cropping, rotation, and color enhancement. Each image is resized to $384 \times 384$ for both training and inference, and the predicted saliency map is bilinearly resized to the original resolution before evaluation. MS-SSIM is evaluated at the same resolution. The backbone is initialized with ImageNet-pretrained SwinV2 weights. The Local Mamba window size is set to $8\times8$. The loss weights are set to $\lambda_s=1.20$ and $\lambda_b=0.10$. We train the model for 70 epochs on two NVIDIA RTX 4090 GPUs using stochastic gradient descent with Nesterov momentum 0.9, weight decay $5 \times 10^{-4}$, gradient clipping 0.5, and a batch size of 8. The initial learning rate is 0.015 for newly introduced layers and is reduced by a factor of 0.1 for the backbone; both learning rates follow a triangular schedule over the training epochs.

\subsection{Quantitative Comparison}

\input{tables/ors4199_challenge_compact.tex}

\begin{figure}[!t]
\centering
\includegraphics[width=0.88\columnwidth]{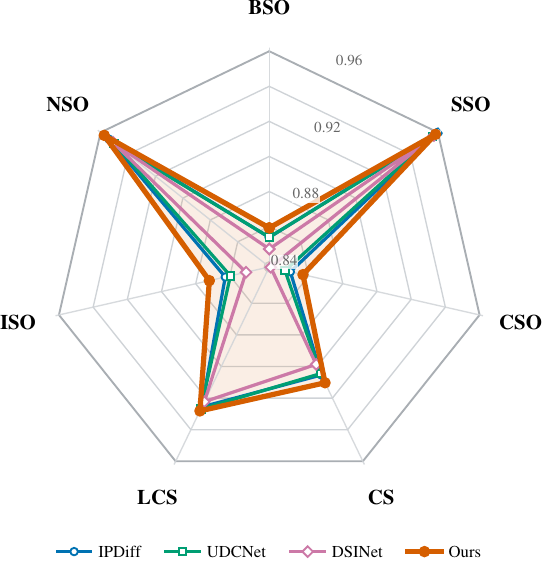}
\caption{MS-SSIM comparison on ORSI-4199 attributes.}
\label{fig:attribute-radar}
\end{figure}

\textbf{Overall comparison.}
We compare SPLG-Mamba with representative ORSI-SOD methods. Table~\ref{tab:overall-sota} summarizes the results on ORSSD and EORSSD. For external methods, we recompute the reported metrics from complete prediction maps released by the original authors using the same evaluation code on the official test sets. Across the two datasets, SPLG-Mamba achieves the best results on most metrics. On ORSSD, SPLG-Mamba ranks first on five of the six metrics. On EORSSD, SPLG-Mamba ranks first on $\mathcal{M}$, $F_\beta^w$, $F_\beta^{\max}$, $E_\xi^{\max}$, and $\overline{F_\beta}$, while remaining close to the best $S_m$. Together, these results indicate improved region structure and threshold-stable foreground separation, which are directly related to complete salient-region prediction.

Figure~\ref{fig:prf} plots threshold-wise precision-recall and F-measure curves on ORSSD, EORSSD, and ORSI-4199. The curves of SPLG-Mamba remain consistently competitive over the threshold range, showing that the advantage in Table~\ref{tab:overall-sota} does not come from a single favorable binarization threshold but from more stable saliency responses.

\textbf{Complexity analysis.}
Table~\ref{tab:complexity} reports the backbone, input size, parameter count, FLOPs, and FPS of representative ORSI-SOD models. The decoder applies Local Mamba to shallow windows, Global Mamba to compact deep features, and lightweight GCSF during cross-scale detail injection, keeping the additional modeling cost concentrated by feature level. Under our experimental settings, SPLG-Mamba reaches 61 FPS with 54.67 G FLOPs. FPS values reported by prior papers are listed for reference because hardware, input size, and implementation details may differ.

\textbf{Comparison on ORSI-4199.} ORSI-4199 provides nine attributes for evaluating performance under different challenging conditions beyond the overall test set. Table~\ref{tab:ors4199-challenge} reports both attribute-based and overall results. On the complete ORSI-4199 test set, SPLG-Mamba ranks first on six of the seven reported metrics and second on $S_m$. The margins are especially clear on MS-SSIM, $F_\beta^w$, and $\overline{F_\beta}$, where SPLG-Mamba exceeds the best competing results by 0.16, 0.90, and 0.81 percentage points, respectively. Taken together, these results indicate higher similarity to the ground-truth maps across multiple scales and better foreground prediction across spatial locations and thresholds.

The attribute-based results further relate the reported measures to the two structural properties defined above. The NSO results provide evidence for continuity in narrow objects, while the ISO and CSO results provide evidence for foreground completeness in incomplete or structurally complex objects. SPLG-Mamba ranks first across all seven metrics on these three attributes. On NSO, SPLG-Mamba improves $F_\beta^w$ and $\overline{F_\beta}$ by 1.46 and 1.26 percentage points, indicating more continuous responses for elongated objects. On ISO, the margins in $S_m$ and $F_\beta^w$ reach 1.03 and 1.10 percentage points, indicating more complete visible-region recovery. On CSO, the gains in $S_m$ and $F_\beta^w$ reach 0.65 and 0.72 percentage points, showing better preservation of internal object structure under complex layouts.

The gains are not restricted to the three structure-focused attributes. On BSO, SPLG-Mamba ranks first on six metrics, and on CS and LCS, it ranks first across all seven metrics. These results indicate that SPLG-Mamba retains region-level consistency for large objects and remains robust when cluttered or weak-contrast backgrounds introduce salient-like distractors. Overall, the ORSI-4199 comparison supports the main claim that SPLG-Mamba improves structural completeness and continuity while maintaining strong localization accuracy across challenging remote-sensing scenes.

Figure~\ref{fig:attribute-radar} complements Table~\ref{tab:ors4199-challenge} by comparing MS-SSIM across seven ORSI-4199 attributes related to object scale (BSO and SSO), object structure and completeness (CSO, NSO, and ISO), and scene complexity and contrast (CS and LCS). SPLG-Mamba achieves the highest MS-SSIM on six of the seven attributes, indicating higher similarity to the ground-truth maps across multiple scales under variations in object scale, object structure, scene complexity, and contrast.

\begin{figure*}[!t]
\centering
\includegraphics[width=\textwidth,height=0.76\textheight,keepaspectratio]{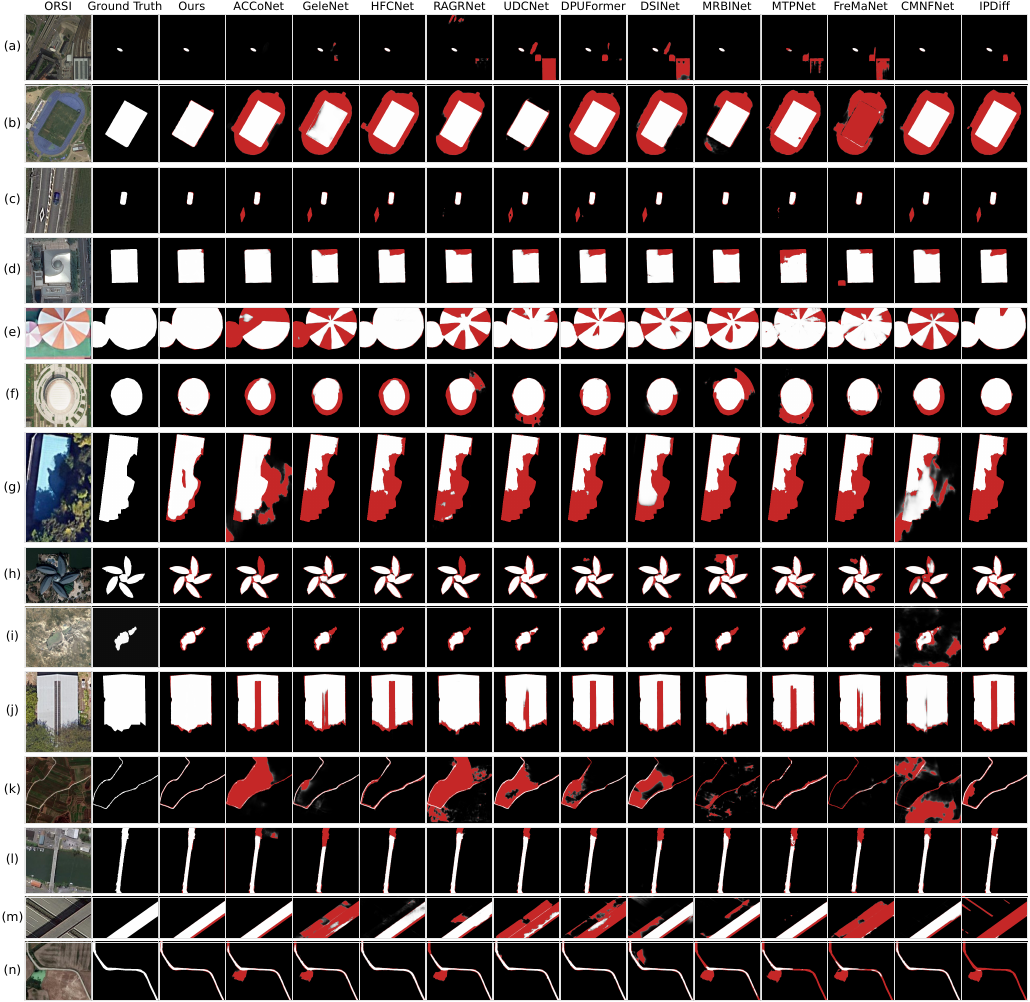}
\caption{Qualitative comparison on ORSI-4199. Red indicates prediction errors.}
\label{fig:qualitative}
\end{figure*}

\subsection{Qualitative Comparison}

Figure~\ref{fig:qualitative} compares SPLG-Mamba with 12 representative competing methods listed in Table~\ref{tab:ors4199-challenge} across 14 ORSI-4199 scenes. Across these examples, SPLG-Mamba more consistently preserves foreground coverage and the continuity of elongated structures while producing fewer false positives on structured backgrounds.

In Fig.~\ref{fig:qualitative}(a)--(c), several competing methods include nearby background structures in their predictions. The resulting errors appear as scattered false positives around the small targets in Fig.~\ref{fig:qualitative}(a) and (c), and as boundary overextension in Fig.~\ref{fig:qualitative}(b). SPLG-Mamba retains the salient regions with fewer background false positives in these scenes.

The broad, structurally complex, or partially visible objects in Fig.~\ref{fig:qualitative}(d)--(g) reveal clearer differences in foreground coverage. Several competing predictions miss visible object regions in Fig.~\ref{fig:qualitative}(d), contain internal gaps or extend beyond the object boundaries in Fig.~\ref{fig:qualitative}(e) and (f), and preserve only part of the visible foreground or include large background regions in Fig.~\ref{fig:qualitative}(g). SPLG-Mamba maintains more complete foreground regions while limiting background inclusion across these examples.

Figures~\ref{fig:qualitative}(h) and (i) show scenes with several spatially separated salient regions. In Fig.~\ref{fig:qualitative}(h), some competing methods omit portions of the four architectural regions or activate nearby background. In Fig.~\ref{fig:qualitative}(i), most competing methods retain the main water region but miss the smaller separated component. SPLG-Mamba preserves the separated foreground regions more completely in both scenes.

The elongated objects in Fig.~\ref{fig:qualitative}(j)--(n) expose differences in structural continuity. Some competing predictions omit narrow components in Fig.~\ref{fig:qualitative}(j) or fragment the irregular object in Fig.~\ref{fig:qualitative}(k). In Fig.~\ref{fig:qualitative}(l)--(n), local breaks and false positives on adjacent linear structures occur in several competing predictions. SPLG-Mamba follows the thin and irregular object paths more continuously while introducing fewer errors in the surrounding background. Taken together, these examples show that SPLG-Mamba better preserves foreground completeness and structural continuity while reducing background false positives.

\input{tables/ablation.tex}

\input{tables/scope_ablation.tex}

\begin{figure*}[!t]
\centering
\includegraphics[width=0.94\textwidth]{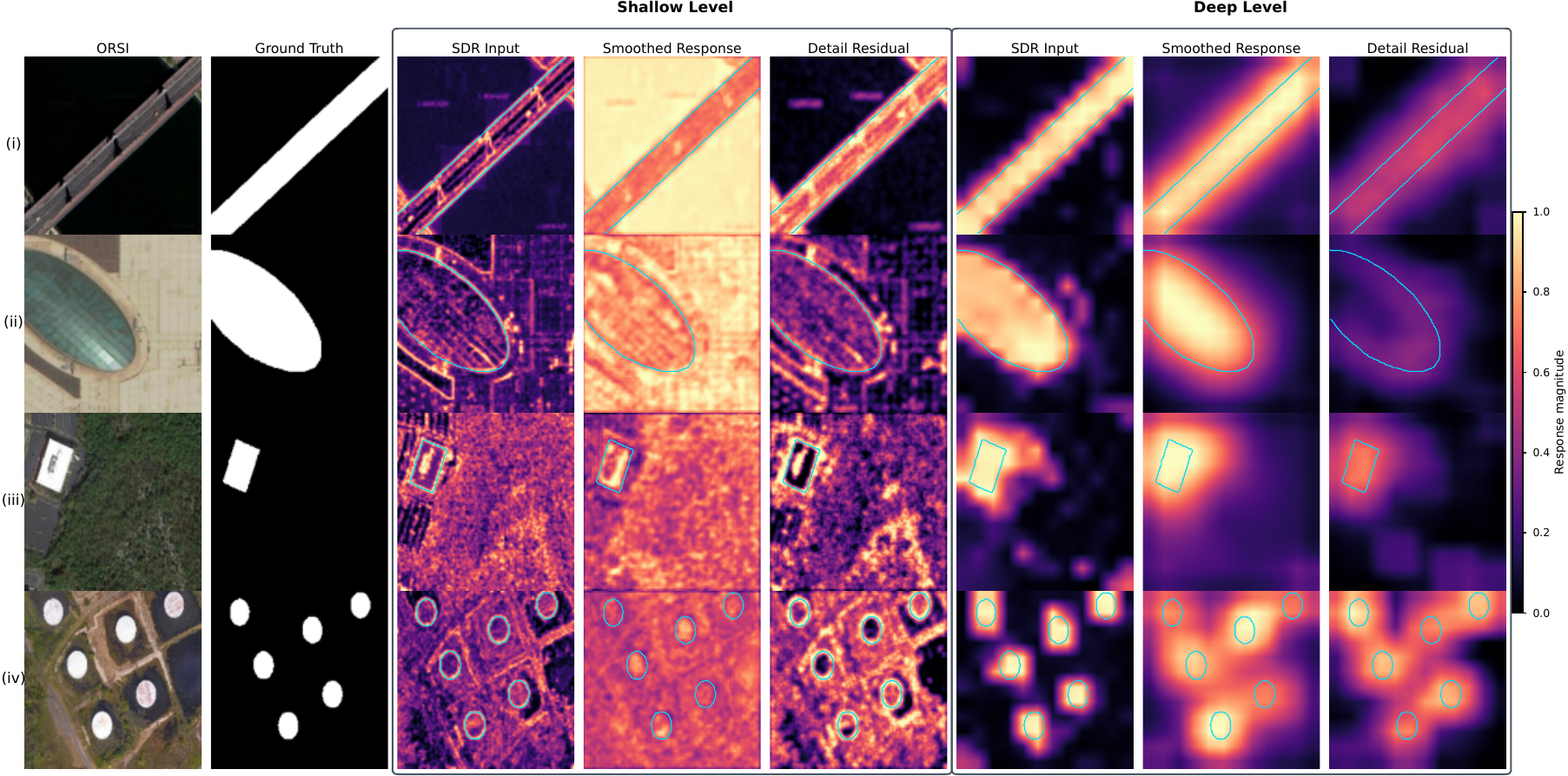}
\caption{SDR input, smoothed response, and detail residual at shallow and deep encoder levels. Cyan indicates ground-truth boundaries. Smooth--detail pairs share a color scale.}
\label{fig:sdr-responses}
\end{figure*}

\begin{figure*}[!t]
\centering
\includegraphics[width=0.94\textwidth]{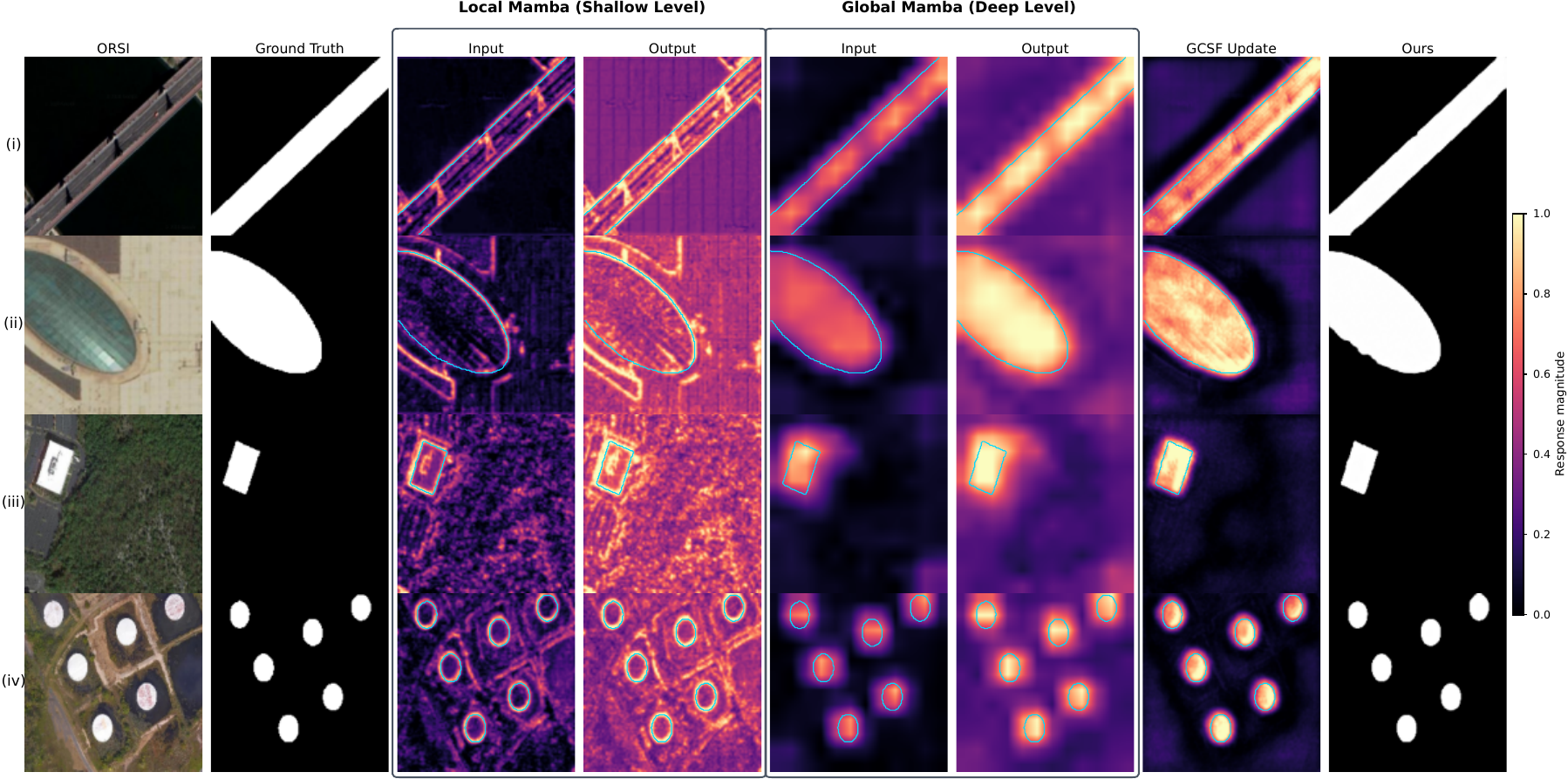}
\caption{Feature responses of Local and Global Mamba and the final GCSF update. Cyan indicates ground-truth boundaries. Each Mamba input--output pair shares a color scale, while the GCSF update is normalized separately.}
\label{fig:mamba-responses}
\end{figure*}

\subsection{Ablation Study}

Table~\ref{tab:ablation} examines how each design choice contributes to the proposed model. All architectural and loss variants are trained and evaluated under the same setting. For the backbone variants, each encoder is initialized with its corresponding pretrained weights, while the remaining architecture, objective, input size, and dataset splits are unchanged. The complete model obtains the best or tied-best result on all six metrics.

\textbf{Model variants.}
Removing SDR degrades performance. Retaining only the smoothed response or only the detail residual reduces $\overline{F_\beta}$ by 1.00 and 1.58 percentage points, respectively. This result indicates the complementary roles of the two components. Replacing GCSF with additive top-down fusion reduces $F_\beta^w$ and $\overline{F_\beta}$ by 0.23 and 0.26 percentage points, respectively. In this variant, the upsampled top-down feature and lateral feature are directly added before convolutional refinement. This result supports the role of GCSF in controlling lateral detail injection into the top-down semantic stream.

The w/o Local Mamba and w/o Global Mamba variants remove local Mamba modeling from shallow features and global Mamba modeling from deep features, respectively, and both lower the F-measure scores. All-local applies windowed Local Mamba at all four levels, whereas All-global applies full-map Global Mamba throughout the hierarchy. Both variants retain Mamba at all levels but use a single spatial scope, producing larger decreases. Replacing Mamba with MHSA under the same spatial scopes produces the largest degradation, reducing $F_\beta^w$ and $\overline{F_\beta}$ by 1.27 and 2.03 percentage points, respectively. These results support both state-space modeling and the hierarchy-specific assignment of local and global Mamba modeling.

Table~\ref{tab:scope-ablation} further analyzes local and global Mamba modeling across object scales and structures. BSO and SSO represent large and small salient objects, while NSO, ISO, and CSO represent narrow, incomplete, and structurally complex salient objects, respectively. The proposed assignment gives the strongest results across all three metrics on BSO, NSO, ISO, and CSO. On SSO, All-local is marginally higher in $S_m$ and MS-SSIM, while the proposed assignment retains the highest $F_\beta^w$. The BSO results support global modeling at deep levels for coherent large-object responses, whereas the SSO results show the contribution of windowed local modeling to small-object features. The consistent advantage on NSO, ISO, and CSO further supports combining local structural modeling with global region-level modeling.

\textbf{Loss and backbone variants.}
Removing any loss term reduces performance without changing the inference architecture. The largest decrease occurs without SSIM, supporting its role in local structural supervision. Removing IoU or Dice also lowers the reported metrics, indicating that both losses provide complementary region-level supervision. The w/o Body Loss variant removes only the auxiliary body loss, and its performance decrease shows that body-mask supervision helps optimize foreground-interior consistency. The backbone variants retain the remaining SPLG-Mamba components, objective, input size, and dataset splits. PVTv2-B5 remains close to SwinV2 across the reported metrics, whereas ResNet-50 shows a larger decrease. This comparison suggests that the current SPLG configuration benefits more from hierarchical Transformer features.

\subsection{Feature Visualization}

Figures~\ref{fig:sdr-responses} and~\ref{fig:mamba-responses} show intermediate feature responses from SDR, Local-Global Mamba, and GCSF. For visualization, absolute feature values are averaged across channels. Figure~\ref{fig:sdr-responses} shows distinct spatial patterns in the two components extracted by SDR. At shallow levels, the detail residual retains fine variations along object boundaries and thin structures while also responding to repeated background textures. The smoothed response exhibits a broader spatial distribution. At deep levels, the detail residual becomes spatially coarser and represents local variations within the semantic feature, whereas the smoothed response retains broader region-level patterns. The coexistence of structural and texture responses in the shallow detail residual motivates its learned comparison with the smoothed response. Based on this comparison, SDR produces a correction that coordinates region-level cues and local structure cues before Mamba modeling. These complementary response patterns are consistent with the ablation results in Table~\ref{tab:ablation}.

The Mamba inputs are channel-projected SDR features, and the paired responses in Fig.~\ref{fig:mamba-responses} show their spatial changes after Mamba modeling. Global Mamba increases target-aligned responses more strongly than the surrounding background. In case (ii), its output becomes more uniform over the entire large object, including the upper region that is weakly activated at the input. In case (iv), Global Mamba maintains coherent activation across multiple spatially separated objects. Local Mamba exhibits a different response pattern at shallow levels. In case (i), the elongated response along the bridge becomes more continuous. In case (iii), the compact rectangular target response remains spatially coherent, whereas nearby vegetation responses remain fragmented. In case (iv), the responses of individual small objects remain distinct from nearby texture patterns. Across the four cases, the GCSF updates are concentrated around foreground boundaries and interiors, and the diffuse background responses visible in the intermediate features are reduced in the final predictions. Together, these patterns provide visual evidence for region-level consistency, local structural continuity, and controlled cross-scale detail injection.

\begin{figure}[t]
\centering
\includegraphics[width=\columnwidth]{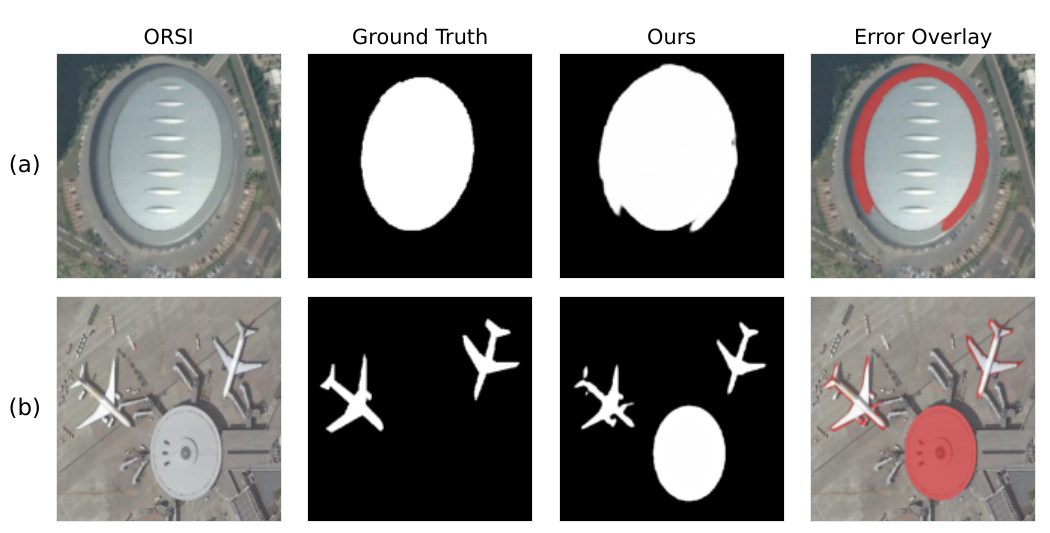}
\caption{Representative failure cases. Red overlays indicate false-positive and false-negative regions.}
\label{fig:failure-cases}
\end{figure}

\subsection{Failure Cases}

The quantitative and qualitative results show that SPLG-Mamba improves foreground completeness and structural continuity across the evaluated datasets. Figure~\ref{fig:failure-cases} presents two challenging cases in which local prediction errors remain. In Fig.~\ref{fig:failure-cases}(a), when salient-object boundaries have appearances similar to adjacent man-made structures, the prediction may slightly overflow around object borders. In Fig.~\ref{fig:failure-cases}(b), in cluttered multi-object scenes with strong man-made backgrounds, the model may preserve the dominant targets while still activating nearby distractors. These cases identify ambiguous boundaries and strong distractors in multi-object scenes as conditions that remain challenging.

\section{Conclusion}

We presented SPLG-Mamba, a Structure-Preserving Local-Global Mamba Network for ORSI-SOD. It addresses structural degradation during hierarchical feature propagation by recalibrating smoothed responses and detail residuals with SDR, assigning local modeling to shallow features and global modeling to deep features through Local-Global Mamba, and controlling cross-scale detail injection with GCSF. Experiments on ORSSD, EORSSD, and ORSI-4199 demonstrate state-of-the-art performance. Attribute-based results further show strong performance for large, complex, narrow, and incomplete objects and in cluttered or low-contrast scenes. Qualitative comparisons, ablation studies, and feature visualization provide additional evidence of improved foreground completeness and structural continuity. Future work will investigate more precise boundary discrimination when salient objects resemble adjacent structures and stronger target separation in cluttered multi-object scenes.

\bibliographystyle{IEEEtran}
\bibliography{references_0601}

\begin{IEEEbiography}[{\includegraphics[width=1in,height=1.25in,clip,keepaspectratio]{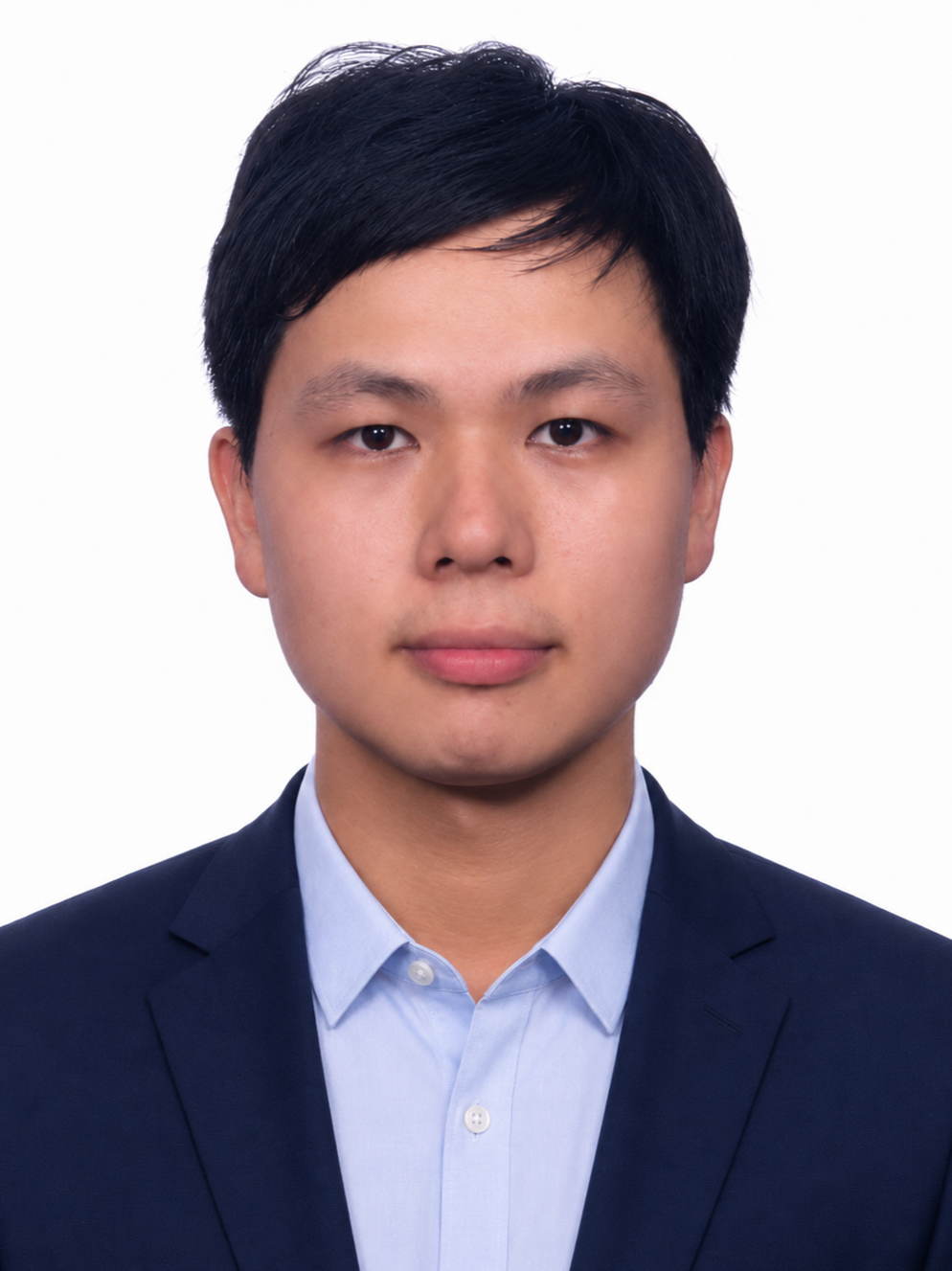}}]{Yi Xu}
received the M.S. degree from Nanjing University, Nanjing, China, in 2024. He is currently pursuing the Ph.D. degree with the School of Software, Nanjing University, Nanjing, China. His research interests include remote sensing image processing, salient object detection, and multimodal learning.
\end{IEEEbiography}

\begin{IEEEbiography}[{\includegraphics[width=1in,height=1.25in,clip,keepaspectratio]{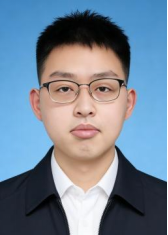}}]{Ruichao Hou}
(Member, IEEE) received his Ph.D. degree from the Department of Computer Science and Technology, Nanjing University, in 2023. He is currently an Associate Professor with the School of Elite Biomedical Engineers and the Institute for Interdisciplinary Intelligent Pharmacy, China Pharmaceutical University. His research mainly focuses on multi-modal representation learning. He has published more than 40 papers in top-tier journals and conferences. He also serves as a reviewer for several prestigious journals, such as IEEE TPAMI, IEEE TNNLS, IEEE TMM, and IEEE TCSVT.
\end{IEEEbiography}

\begin{IEEEbiography}[{\includegraphics[width=0.9in,height=1.25in,clip,keepaspectratio]{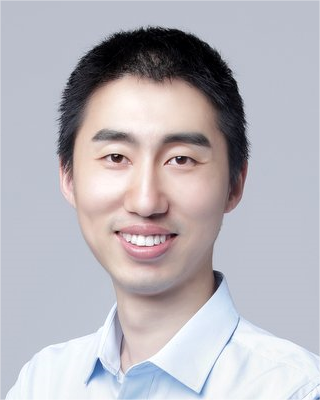}}]{Tongwei Ren}
(Member, IEEE) received the B.S., M.E., and Ph.D. degrees from Nanjing University, Nanjing, China, in 2004, 2006, and 2010, respectively. He joined Nanjing University in 2010, and at present he is a professor. His research interest mainly includes multimedia computing and its real-world applications. He has published more than 40 papers in top-tier journals and conferences. He was a recipient of the best paper candidate awards of ICIMCS 2014, PCM 2015, and MMAsia 2020, and he was in the champion teams of ECCV 2018 PIC challenge, MM 2019 VRU challenge, MM 2020 DVU challenge, MM 2022 DVU challenge, and MM 2023 DVU challenge.
\end{IEEEbiography}

\begin{IEEEbiography}[{\includegraphics[width=1in,height=1.25in,clip,keepaspectratio]{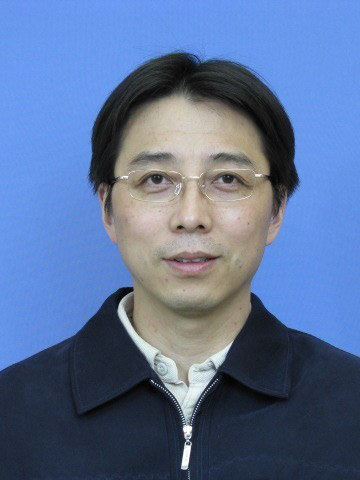}}]{Gangshan Wu}
(Member, IEEE) received the B.Sc., M.S., and Ph.D. degrees from the Department of Computer Science and Technology, Nanjing University, Nanjing, China, in 1988, 1991, and 2000, respectively. He is currently a Professor with the School of Computer Science, Nanjing University. His current research interests include computer vision, multimedia content analysis, multimedia information retrieval, digital museum, and large-scale volumetric data processing.
\end{IEEEbiography}

\end{document}

%% file: tables/overall_comparison.tex
\begin{table*}[!t]
\centering
\caption{Quantitative comparison on ORSSD and EORSSD. \textcolor{red}{Red}, \textcolor{green!60!black}{green}, and \textcolor{blue}{blue} highlight the first, second, and third best results, respectively; -- denotes unavailable results.}
\label{tab:overall-sota}
\scriptsize
\setlength{\tabcolsep}{1.65pt}
\resizebox{\textwidth}{!}{%
\begin{tabular}{llcccccccccccc}
\toprule
\multirow{2}{*}{Method} & \multirow{2}{*}{Pub.} & \multicolumn{6}{c}{ORSSD} & \multicolumn{6}{c}{EORSSD} \\
\cmidrule(lr){3-8}\cmidrule(lr){9-14}
 & & $\mathcal{M}\downarrow$ & $S_m\uparrow$ & $F_\beta^w\uparrow$ & $F_\beta^{\max}\uparrow$ & $E_\xi^{\max}\uparrow$ & $\overline{F_\beta}\uparrow$ & $\mathcal{M}\downarrow$ & $S_m\uparrow$ & $F_\beta^w\uparrow$ & $F_\beta^{\max}\uparrow$ & $E_\xi^{\max}\uparrow$ & $\overline{F_\beta}\uparrow$ \\
\midrule
DCPNet~\cite{li2024dcpnet} & Electron.,2024 & .0073 & .9498 & .9140 & .9326 & .9855 & .9124 & .0053 & .9409 & .8869 & .9074 & .9817 & .8812 \\
HFCNet~\cite{liu2024hfcnet} & TGRS,2024 & .0070 & .9543 & .9079 & .9349 & .9884 & .9049 & .0049 & \textcolor{green!60!black}{.9449} & .8698 & .9055 & .9842 & .8646 \\
RAGRNet~\cite{zhao2024ragrnet} & TGRS,2024 & .0066 & .9507 & .9157 & .9311 & .9861 & .9157 & .0057 & .9364 & .8774 & .8951 & .9785 & .8770 \\
UDCNet~\cite{sun2024udcnet} & TGRS,2024 & \textcolor{red}{.0052} & \textcolor{blue}{.9553} & .9204 & .9354 & .9900 & .9145 & .0050 & .9385 & .8778 & .9014 & .9819 & .8649 \\
DPUFormer~\cite{sun2025dpuformer} & IJCAI,2025 & .0062 & .9412 & .9155 & .9353 & .9868 & .9142 & .0056 & .9401 & .8809 & .9051 & .9816 & .8757 \\
DSINet~\cite{ge2025enhanced} & VC,2025 & .0080 & .9507 & .9108 & .9340 & .9881 & .9082 & .0056 & .9417 & .8826 & .9065 & .9841 & .8776 \\
LightEMNet~\cite{xing2025lightemnet} & TGRS,2025 & -- & -- & -- & -- & -- & -- & .0081 & .8872 & .8203 & .8683 & .9628 & .8351 \\
MRBINet~\cite{jia2025mrbinet} & TGRS,2025 & .0069 & .9474 & .9110 & .9271 & .9851 & .9110 & .0056 & .9354 & .8759 & .8960 & .9766 & .8768 \\
MTPNet~\cite{luo2025mtpnet} & JSTARS,2025 & .0066 & .9538 & \textcolor{blue}{.9230} & \textcolor{blue}{.9379} & \textcolor{green!60!black}{.9916} & \textcolor{blue}{.9195} & .0050 & .9387 & .8837 & .9025 & .9808 & .8797 \\
RAMENet~\cite{han2025ramenet} & TGRS,2025 & .0072 & .9489 & .9137 & .9322 & .9862 & .9147 & .0050 & .9420 & .8868 & .9071 & .9822 & \textcolor{blue}{.8840} \\
THMNet~\cite{yang2025thmnet} & TGRS,2025 & .0071 & .9506 & .9140 & .9327 & .9843 & .9124 & .0054 & .9431 & \textcolor{blue}{.8885} & \textcolor{blue}{.9092} & .9826 & .8820 \\
CMNFNet~\cite{xu2025cmnfnet} & TCYB,2025 & .0078 & .9475 & .9017 & .9279 & .9832 & .9029 & .0063 & .9378 & .8632 & .8977 & .9774 & .8590 \\
FreMaNet~\cite{li2026frema} & TGRS,2026 & .0067 & .9496 & .9126 & .9300 & .9858 & .9118 & \textcolor{blue}{.0048} & .9428 & \textcolor{green!60!black}{.8899} & .9089 & \textcolor{blue}{.9844} & \textcolor{green!60!black}{.8864} \\
IPDiff~\cite{li2026ipdiff} & IJCV,2026 & \textcolor{blue}{.0054} & \textcolor{green!60!black}{.9557} & \textcolor{green!60!black}{.9247} & \textcolor{green!60!black}{.9412} & \textcolor{blue}{.9915} & \textcolor{green!60!black}{.9200} & \textcolor{green!60!black}{.0044} & \textcolor{red}{.9462} & .8841 & \textcolor{green!60!black}{.9108} & \textcolor{green!60!black}{.9861} & .8745 \\
\midrule
Ours & -- & \textcolor{green!60!black}{.0053} & \textcolor{red}{.9564} & \textcolor{red}{.9290} & \textcolor{red}{.9433} & \textcolor{red}{.9924} & \textcolor{red}{.9250} & \textcolor{red}{.0041} & \textcolor{blue}{.9447} & \textcolor{red}{.8949} & \textcolor{red}{.9113} & \textcolor{red}{.9865} & \textcolor{red}{.8877} \\
\bottomrule
\end{tabular}}
\end{table*}

%% file: tables/complexity.tex
\begin{table}[!t]
\centering
\caption{Comparison of model complexity and inference speed.}
\label{tab:complexity}
\scriptsize
\setlength{\tabcolsep}{1.7pt}
\resizebox{\columnwidth}{!}{%
\begin{tabular}{lllccc}
\toprule
Method & Backbone & Input & Params (M) $\downarrow$ & FLOPs (G) $\downarrow$ & FPS $\uparrow$ \\
\midrule
DCPNet~\cite{li2024dcpnet} & PVTv2+ResNet & $352\times352$ & 99.31 & 20.52 & -- \\
HFCNet~\cite{liu2024hfcnet} & VGG+Swin & $224\times224$ & 140.75 & 120.41 & 38 \\
RAGRNet~\cite{zhao2024ragrnet} & Res2Net & $256\times256$ & 35.60 & 17.82 & 36 \\
UDCNet~\cite{sun2024udcnet} & ResNet & $352\times352$ & 72.20 & 101.19 & -- \\
DPUFormer~\cite{sun2025dpuformer} & DPU-Former & $352\times352$ & 44.20 & 32.51 & -- \\
DSINet~\cite{ge2025enhanced} & CNN--Trans. & $352\times352$ & 109.89 & 89.69 & -- \\
LightEMNet~\cite{xing2025lightemnet} & MobileNetV2 & $352\times352$ & 4.81 & 23.03 & -- \\
MRBINet~\cite{jia2025mrbinet} & Res2Net & $256\times256$ & 32.40 & 42.80 & 9 \\
MTPNet~\cite{luo2025mtpnet} & Swin-Tiny & $352\times352$ & 55.06 & -- & 91 \\
RAMENet~\cite{han2025ramenet} & MobileViT & $352\times352$ & 5.18 & 8.72 & 92 \\
THMNet~\cite{yang2025thmnet} & PVT & $352\times352$ & 33.72 & 13.02 & 74 \\
FreMaNet~\cite{li2026frema} & MobileViT & $352\times352$ & 4.91 & 4.52 & 208 \\
IPDiff~\cite{li2026ipdiff} & PVT & $352\times352$ & 82.60 & 70.40 & 4 \\
\midrule
Ours & Swin & $384\times384$ & 107.27 & 54.67 & 61 \\
\bottomrule
\end{tabular}}
\end{table}

%% file: tables/ors4199_challenge_compact.tex
\begin{table*}[!t]
\centering
\caption{Quantitative comparison on the challenging ORSI-4199 dataset. \textcolor{red}{Red}, \textcolor{green!60!black}{green}, and \textcolor{blue}{blue} highlight the first, second, and third best results, respectively.}
\label{tab:ors4199-challenge}
\scriptsize
\setlength{\tabcolsep}{1.3pt}
\renewcommand{\arraystretch}{0.70}
\newcommand{\orsmetric}[1]{{\tiny $#1$}}
\resizebox{0.85\textwidth}{!}{%
\begin{tabular}{llccccccccccc}
\toprule
Attribute & Metric & HFCNet & RAGRNet & UDCNet & DPUFormer & DSINet & MRBINet & MTPNet & FreMaNet & CMNFNet & IPDiff & Ours \\
\midrule
\multirow{7}{*}{BSO} & \orsmetric{\mathcal{M}\downarrow} & .0613 & .0593 & \textcolor{green!60!black}{.0473} & .0555 & .0532 & .0559 & .0587 & .0565 & .0606 & \textcolor{blue}{.0491} & \textcolor{red}{.0452}\\
 & \orsmetric{S_m\uparrow} & .8881 & .8868 & \textcolor{green!60!black}{.9055} & .8913 & .9003 & .8909 & .8877 & .8885 & .8880 & \textcolor{blue}{.9022} & \textcolor{red}{.9083}\\
 & \orsmetric{\mathrm{MS\!-\!SSIM}\uparrow} & .8377 & .8402 & \textcolor{green!60!black}{.8540} & .8405 & .8472 & .8436 & .8352 & .8320 & .8360 & \textcolor{blue}{.8538} & \textcolor{red}{.8593}\\
 & \orsmetric{F_\beta^w\uparrow} & .8907 & .8956 & \textcolor{green!60!black}{.9194} & .9029 & .9080 & .9003 & .8966 & .8972 & .8925 & \textcolor{blue}{.9128} & \textcolor{red}{.9218}\\
 & \orsmetric{F_\beta^{\max}\uparrow} & .9375 & .9372 & \textcolor{red}{.9511} & .9411 & \textcolor{blue}{.9480} & .9372 & .9404 & .9439 & .9383 & .9432 & \textcolor{green!60!black}{.9502}\\
 & \orsmetric{E_\xi^{\max}\uparrow} & .9231 & .9249 & \textcolor{green!60!black}{.9404} & .9304 & \textcolor{blue}{.9366} & .9285 & .9286 & .9328 & .9266 & .9349 & \textcolor{red}{.9442}\\
 & \orsmetric{\overline{F_\beta}\uparrow} & .9128 & .9217 & \textcolor{green!60!black}{.9375} & .9268 & .9283 & .9244 & .9203 & .9214 & .9158 & \textcolor{blue}{.9325} & \textcolor{red}{.9401}\\
\hline
\multirow{7}{*}{SSO} & \orsmetric{\mathcal{M}\downarrow} & .0075 & .0079 & .0080 & .0072 & .0075 & .0089 & \textcolor{green!60!black}{.0068} & .0076 & .0110 & \textcolor{red}{.0067} & \textcolor{blue}{.0071}\\
 & \orsmetric{S_m\uparrow} & .8536 & .8553 & .8522 & .8554 & .8515 & .8527 & \textcolor{green!60!black}{.8654} & \textcolor{blue}{.8556} & .8443 & \textcolor{red}{.8655} & .8520\\
 & \orsmetric{\mathrm{MS\!-\!SSIM}\uparrow} & .9568 & .9545 & .9562 & .9565 & \textcolor{blue}{.9571} & .9497 & .9554 & .9531 & .9487 & \textcolor{red}{.9604} & \textcolor{green!60!black}{.9583}\\
 & \orsmetric{F_\beta^w\uparrow} & .8063 & .8088 & .8079 & \textcolor{green!60!black}{.8212} & .8095 & .7987 & \textcolor{blue}{.8203} & .8120 & .7906 & .8173 & \textcolor{red}{.8222}\\
 & \orsmetric{F_\beta^{\max}\uparrow} & .8444 & .8394 & .8353 & \textcolor{green!60!black}{.8496} & .8471 & .8313 & .8472 & \textcolor{blue}{.8479} & .8320 & .8443 & \textcolor{red}{.8520}\\
 & \orsmetric{E_\xi^{\max}\uparrow} & .9621 & .9541 & .9582 & .9618 & .9580 & .9464 & \textcolor{blue}{.9627} & .9588 & .9506 & \textcolor{red}{.9634} & \textcolor{green!60!black}{.9630}\\
 & \orsmetric{\overline{F_\beta}\uparrow} & .8146 & .8180 & .8065 & \textcolor{blue}{.8262} & .8151 & .8104 & \textcolor{green!60!black}{.8275} & .8218 & .7992 & .8215 & \textcolor{red}{.8281}\\
\hline
\multirow{7}{*}{OC} & \orsmetric{\mathcal{M}\downarrow} & .0151 & .0146 & .0130 & .0118 & \textcolor{green!60!black}{.0113} & .0138 & .0118 & \textcolor{red}{.0107} & .0135 & .0128 & \textcolor{blue}{.0114}\\
 & \orsmetric{S_m\uparrow} & .8609 & .8633 & .8626 & .8667 & .8666 & .8618 & \textcolor{red}{.8785} & \textcolor{blue}{.8742} & .8600 & \textcolor{green!60!black}{.8753} & .8669\\
 & \orsmetric{\mathrm{MS\!-\!SSIM}\uparrow} & .9560 & .9545 & .9565 & .9592 & \textcolor{red}{.9625} & .9513 & .9597 & .9610 & .9553 & \textcolor{blue}{.9611} & \textcolor{green!60!black}{.9623}\\
 & \orsmetric{F_\beta^w\uparrow} & .8292 & .8311 & .8381 & .8492 & .8416 & .8225 & \textcolor{blue}{.8493} & \textcolor{green!60!black}{.8500} & .8255 & .8409 & \textcolor{red}{.8523}\\
 & \orsmetric{F_\beta^{\max}\uparrow} & .8670 & .8638 & .8685 & .8790 & .8757 & .8570 & \textcolor{blue}{.8793} & \textcolor{red}{.8831} & .8649 & .8687 & \textcolor{green!60!black}{.8811}\\
 & \orsmetric{E_\xi^{\max}\uparrow} & .9567 & .9511 & .9592 & \textcolor{green!60!black}{.9639} & .9622 & .9429 & \textcolor{blue}{.9631} & \textcolor{red}{.9670} & .9559 & .9576 & .9614\\
 & \orsmetric{\overline{F_\beta}\uparrow} & .8376 & .8430 & .8395 & .8565 & .8485 & .8360 & \textcolor{blue}{.8586} & \textcolor{red}{.8602} & .8342 & .8460 & \textcolor{red}{.8602}\\
\hline
\multirow{7}{*}{CSO} & \orsmetric{\mathcal{M}\downarrow} & .0668 & .0661 & \textcolor{green!60!black}{.0532} & .0614 & .0588 & .0618 & .0648 & .0593 & .0653 & \textcolor{blue}{.0536} & \textcolor{red}{.0486}\\
 & \orsmetric{S_m\uparrow} & .8675 & .8630 & \textcolor{green!60!black}{.8852} & .8701 & .8802 & .8701 & .8667 & .8719 & .8689 & \textcolor{blue}{.8827} & \textcolor{red}{.8917}\\
 & \orsmetric{\mathrm{MS\!-\!SSIM}\uparrow} & .8310 & .8285 & \textcolor{blue}{.8462} & .8346 & .8379 & .8352 & .8259 & .8285 & .8297 & \textcolor{green!60!black}{.8493} & \textcolor{red}{.8567}\\
 & \orsmetric{F_\beta^w\uparrow} & .8611 & .8613 & \textcolor{green!60!black}{.8882} & .8722 & .8776 & .8701 & .8642 & .8723 & .8631 & \textcolor{blue}{.8843} & \textcolor{red}{.8954}\\
 & \orsmetric{F_\beta^{\max}\uparrow} & .9114 & .9068 & \textcolor{green!60!black}{.9222} & .9120 & .9204 & .9091 & .9102 & \textcolor{blue}{.9206} & .9108 & .9152 & \textcolor{red}{.9235}\\
 & \orsmetric{E_\xi^{\max}\uparrow} & .9167 & .9152 & \textcolor{green!60!black}{.9331} & .9235 & \textcolor{blue}{.9299} & .9207 & .9192 & .9297 & .9202 & .9291 & \textcolor{red}{.9398}\\
 & \orsmetric{\overline{F_\beta}\uparrow} & .8827 & .8890 & \textcolor{green!60!black}{.9060} & .8969 & .8978 & .8949 & .8882 & .8975 & .8863 & \textcolor{blue}{.9039} & \textcolor{red}{.9126}\\
\hline
\multirow{7}{*}{CS} & \orsmetric{\mathcal{M}\downarrow} & .0383 & .0385 & \textcolor{green!60!black}{.0316} & .0357 & .0349 & .0365 & .0366 & .0368 & .0393 & \textcolor{blue}{.0325} & \textcolor{red}{.0293}\\
 & \orsmetric{S_m\uparrow} & .9002 & .8956 & \textcolor{blue}{.9073} & .9000 & .9030 & .8971 & .8996 & .8962 & .8949 & \textcolor{green!60!black}{.9074} & \textcolor{red}{.9126}\\
 & \orsmetric{\mathrm{MS\!-\!SSIM}\uparrow} & .8966 & .8939 & \textcolor{blue}{.9045} & .8964 & .8987 & .8951 & .8935 & .8893 & .8914 & \textcolor{green!60!black}{.9054} & \textcolor{red}{.9103}\\
 & \orsmetric{F_\beta^w\uparrow} & .8886 & .8862 & \textcolor{green!60!black}{.9029} & .8973 & .8925 & .8868 & .8925 & .8876 & .8780 & \textcolor{blue}{.9013} & \textcolor{red}{.9132}\\
 & \orsmetric{F_\beta^{\max}\uparrow} & .9292 & .9227 & \textcolor{green!60!black}{.9327} & .9297 & \textcolor{blue}{.9326} & .9210 & .9275 & .9303 & .9211 & .9288 & \textcolor{red}{.9394}\\
 & \orsmetric{E_\xi^{\max}\uparrow} & .9463 & .9418 & \textcolor{green!60!black}{.9536} & .9488 & .9502 & .9429 & .9490 & .9471 & .9424 & \textcolor{blue}{.9519} & \textcolor{red}{.9602}\\
 & \orsmetric{\overline{F_\beta}\uparrow} & .9047 & .9062 & \textcolor{blue}{.9147} & \textcolor{blue}{.9147} & .9080 & .9058 & .9094 & .9077 & .8956 & \textcolor{green!60!black}{.9157} & \textcolor{red}{.9267}\\
\hline
\multirow{7}{*}{NSO} & \orsmetric{\mathcal{M}\downarrow} & .0180 & .0212 & .0166 & .0170 & .0165 & .0170 & \textcolor{blue}{.0160} & .0211 & .0194 & \textcolor{green!60!black}{.0156} & \textcolor{red}{.0144}\\
 & \orsmetric{S_m\uparrow} & .8998 & .8976 & .8985 & .8981 & .8976 & .8996 & \textcolor{green!60!black}{.9088} & .8909 & .8976 & \textcolor{blue}{.9044} & \textcolor{red}{.9100}\\
 & \orsmetric{\mathrm{MS\!-\!SSIM}\uparrow} & \textcolor{blue}{.9521} & .9401 & .9502 & .9502 & .9509 & .9449 & .9508 & .9368 & .9463 & \textcolor{green!60!black}{.9539} & \textcolor{red}{.9573}\\
 & \orsmetric{F_\beta^w\uparrow} & .8852 & .8779 & .8929 & \textcolor{blue}{.8956} & .8876 & .8815 & \textcolor{green!60!black}{.8972} & .8745 & .8779 & .8952 & \textcolor{red}{.9118}\\
 & \orsmetric{F_\beta^{\max}\uparrow} & .9218 & .9108 & .9216 & .9239 & \textcolor{blue}{.9243} & .9145 & \textcolor{green!60!black}{.9269} & .9138 & .9169 & .9211 & \textcolor{red}{.9340}\\
 & \orsmetric{E_\xi^{\max}\uparrow} & .9685 & .9624 & .9675 & .9701 & .9706 & .9671 & \textcolor{green!60!black}{.9724} & .9643 & .9685 & \textcolor{blue}{.9708} & \textcolor{red}{.9759}\\
 & \orsmetric{\overline{F_\beta}\uparrow} & .8905 & .8905 & .8964 & \textcolor{blue}{.9038} & .8940 & .8939 & \textcolor{green!60!black}{.9043} & .8885 & .8870 & .9004 & \textcolor{red}{.9169}\\
\hline
\multirow{7}{*}{MSO} & \orsmetric{\mathcal{M}\downarrow} & .0167 & .0168 & \textcolor{blue}{.0153} & .0161 & .0166 & .0178 & .0163 & .0162 & .0196 & \textcolor{green!60!black}{.0149} & \textcolor{red}{.0148}\\
 & \orsmetric{S_m\uparrow} & .8808 & .8825 & .8831 & \textcolor{blue}{.8833} & .8780 & .8802 & \textcolor{green!60!black}{.8864} & .8806 & .8718 & \textcolor{red}{.8869} & .8799\\
 & \orsmetric{\mathrm{MS\!-\!SSIM}\uparrow} & \textcolor{green!60!black}{.9419} & .9370 & \textcolor{blue}{.9415} & .9398 & .9400 & .9338 & .9354 & .9349 & .9340 & \textcolor{red}{.9445} & .9414\\
 & \orsmetric{F_\beta^w\uparrow} & .8521 & .8539 & .8554 & \textcolor{red}{.8632} & .8486 & .8454 & .8566 & .8553 & .8378 & \textcolor{blue}{.8618} & \textcolor{green!60!black}{.8621}\\
 & \orsmetric{F_\beta^{\max}\uparrow} & .8860 & .8857 & .8821 & \textcolor{red}{.8906} & .8848 & .8783 & .8839 & \textcolor{green!60!black}{.8881} & .8767 & .8865 & \textcolor{blue}{.8875}\\
 & \orsmetric{E_\xi^{\max}\uparrow} & .9650 & .9598 & \textcolor{blue}{.9654} & \textcolor{blue}{.9654} & .9620 & .9531 & .9636 & .9639 & .9580 & \textcolor{red}{.9664} & \textcolor{green!60!black}{.9659}\\
 & \orsmetric{\overline{F_\beta}\uparrow} & .8593 & .8667 & .8551 & \textcolor{red}{.8710} & .8558 & .8601 & .8664 & \textcolor{blue}{.8671} & .8483 & .8670 & \textcolor{green!60!black}{.8696}\\
\hline
\multirow{7}{*}{LCS} & \orsmetric{\mathcal{M}\downarrow} & .0161 & .0166 & \textcolor{green!60!black}{.0136} & .0151 & .0163 & .0158 & .0167 & .0174 & .0221 & \textcolor{blue}{.0145} & \textcolor{red}{.0134}\\
 & \orsmetric{S_m\uparrow} & .8674 & .8631 & \textcolor{green!60!black}{.8703} & .8679 & .8671 & .8608 & .8663 & .8589 & .8560 & \textcolor{blue}{.8680} & \textcolor{red}{.8739}\\
 & \orsmetric{\mathrm{MS\!-\!SSIM}\uparrow} & .9232 & .9178 & \textcolor{green!60!black}{.9269} & .9225 & .9223 & .9177 & .9185 & .9137 & .9101 & \textcolor{blue}{.9256} & \textcolor{red}{.9281}\\
 & \orsmetric{F_\beta^w\uparrow} & .8009 & .8018 & \textcolor{blue}{.8115} & \textcolor{green!60!black}{.8121} & .8039 & .7971 & .8102 & .7979 & .7845 & .8037 & \textcolor{red}{.8225}\\
 & \orsmetric{F_\beta^{\max}\uparrow} & .8369 & .8328 & .8386 & \textcolor{green!60!black}{.8403} & .8393 & .8273 & \textcolor{blue}{.8396} & .8352 & .8264 & .8288 & \textcolor{red}{.8466}\\
 & \orsmetric{E_\xi^{\max}\uparrow} & .9645 & .9626 & \textcolor{green!60!black}{.9669} & \textcolor{blue}{.9651} & .9632 & .9576 & .9641 & .9606 & .9541 & .9647 & \textcolor{red}{.9722}\\
 & \orsmetric{\overline{F_\beta}\uparrow} & .8108 & .8148 & .8162 & \textcolor{green!60!black}{.8211} & .8128 & .8106 & \textcolor{green!60!black}{.8211} & .8113 & .7972 & .8114 & \textcolor{red}{.8305}\\
\hline
\multirow{7}{*}{ISO} & \orsmetric{\mathcal{M}\downarrow} & .0715 & .0688 & \textcolor{green!60!black}{.0540} & .0669 & .0610 & .0646 & .0645 & .0602 & .0712 & \textcolor{blue}{.0550} & \textcolor{red}{.0491}\\
 & \orsmetric{S_m\uparrow} & .8665 & .8676 & \textcolor{green!60!black}{.8872} & .8687 & .8823 & .8734 & .8731 & .8777 & .8668 & \textcolor{blue}{.8846} & \textcolor{red}{.8975}\\
 & \orsmetric{\mathrm{MS\!-\!SSIM}\uparrow} & .8417 & .8468 & \textcolor{blue}{.8597} & .8429 & .8506 & .8477 & .8453 & .8406 & .8372 & \textcolor{green!60!black}{.8627} & \textcolor{red}{.8720}\\
 & \orsmetric{F_\beta^w\uparrow} & .8755 & .8834 & \textcolor{green!60!black}{.9059} & .8879 & .8944 & .8927 & .8903 & .8942 & .8750 & \textcolor{blue}{.9018} & \textcolor{red}{.9169}\\
 & \orsmetric{F_\beta^{\max}\uparrow} & .9313 & .9318 & \textcolor{green!60!black}{.9423} & .9348 & .9401 & .9350 & .9360 & \textcolor{blue}{.9413} & .9309 & .9361 & \textcolor{red}{.9458}\\
 & \orsmetric{E_\xi^{\max}\uparrow} & .9128 & .9166 & \textcolor{blue}{.9317} & .9219 & .9271 & .9219 & .9277 & \textcolor{green!60!black}{.9318} & .9134 & .9294 & \textcolor{red}{.9401}\\
 & \orsmetric{\overline{F_\beta}\uparrow} & .8990 & .9134 & \textcolor{green!60!black}{.9245} & .9152 & .9159 & .9194 & .9150 & .9195 & .9023 & \textcolor{blue}{.9211} & \textcolor{red}{.9347}\\
\hline
\multirow{7}{*}{Overall} & \orsmetric{\mathcal{M}\downarrow} & .0281 & .0284 & \textcolor{green!60!black}{.0237} & .0263 & .0257 & .0268 & .0268 & .0270 & .0301 & \textcolor{green!60!black}{.0237} & \textcolor{red}{.0223}\\
 & \orsmetric{S_m\uparrow} & .8827 & .8811 & \textcolor{blue}{.8872} & .8833 & .8844 & .8824 & .8862 & .8810 & .8774 & \textcolor{red}{.8907} & \textcolor{green!60!black}{.8901}\\
 & \orsmetric{\mathrm{MS\!-\!SSIM}\uparrow} & .9161 & .9125 & \textcolor{blue}{.9207} & .9158 & .9177 & .9131 & .9130 & .9095 & .9099 & \textcolor{green!60!black}{.9227} & \textcolor{red}{.9243}\\
 & \orsmetric{F_\beta^w\uparrow} & .8596 & .8602 & \textcolor{blue}{.8709} & .8704 & .8644 & .8594 & .8674 & .8615 & .8504 & \textcolor{green!60!black}{.8718} & \textcolor{red}{.8808}\\
 & \orsmetric{F_\beta^{\max}\uparrow} & .8981 & .8942 & .8989 & \textcolor{blue}{.9011} & \textcolor{green!60!black}{.9017} & .8922 & .8998 & .9007 & .8917 & .8983 & \textcolor{red}{.9070}\\
 & \orsmetric{E_\xi^{\max}\uparrow} & .9525 & .9492 & \textcolor{blue}{.9568} & .9547 & .9552 & .9489 & .9553 & .9538 & .9485 & \textcolor{green!60!black}{.9572} & \textcolor{red}{.9615}\\
 & \orsmetric{\overline{F_\beta}\uparrow} & .8718 & .8763 & .8777 & \textcolor{green!60!black}{.8831} & .8754 & .8755 & .8806 & .8775 & .8643 & \textcolor{blue}{.8817} & \textcolor{red}{.8912}\\
\bottomrule
\end{tabular}}
\end{table*}

%% file: tables/ablation.tex
\begin{table}[!t]
\centering
\caption{Ablation study on ORSI-4199. The best result in each column is highlighted in \textcolor{red}{red}.}
\label{tab:ablation}
\scriptsize
\setlength{\tabcolsep}{2.0pt}
\resizebox{\columnwidth}{!}{%
\begin{tabular}{lcccccc}
\toprule
Variant & $\mathcal{M}\downarrow$ & $S_m\uparrow$ & $F_\beta^w\uparrow$ & $F_\beta^{\max}\uparrow$ & $E_\xi^{\max}\uparrow$ & $\overline{F_\beta}\uparrow$ \\
\midrule
\multicolumn{7}{l}{\textit{Model variants}} \\
w/o SDR & .0226 & .8896 & .8803 & .9064 & \textcolor{red}{.9615} & .8894 \\
Smooth only & .0233 & .8870 & .8743 & .9017 & .9591 & .8812 \\
Detail only & .0231 & .8854 & .8706 & .8990 & .9582 & .8754 \\
w/o Local Mamba & .0232 & .8894 & .8803 & .9067 & .9610 & .8896 \\
w/o Global Mamba & .0235 & .8883 & .8794 & .9063 & .9603 & .8896 \\
All-local & .0235 & .8867 & .8748 & .9021 & .9590 & .8823 \\
All-global & .0234 & .8869 & .8743 & .9019 & .9592 & .8817 \\
Mamba $\rightarrow$ MHSA & .0233 & .8849 & .8681 & .8980 & .9584 & .8709 \\
w/o GCSF & .0234 & .8879 & .8785 & .9063 & .9607 & .8886 \\
\midrule
\multicolumn{7}{l}{\textit{Loss variants}} \\
w/o IoU Loss & .0232 & .8874 & .8746 & .9031 & .9593 & .8823 \\
w/o Dice Loss & .0234 & .8871 & .8746 & .9014 & .9596 & .8815 \\
w/o SSIM Loss & .0238 & .8859 & .8724 & .9005 & .9591 & .8771 \\
w/o Body Supervision & .0232 & .8891 & .8794 & .9057 & .9608 & .8885 \\
\midrule
\multicolumn{7}{l}{\textit{Backbone variants}} \\
ResNet-50 & .0348 & .8607 & .8413 & .8802 & .9285 & .8596 \\
PVTv2-B5 & .0258 & .8813 & .8708 & .8951 & .9506 & .8808 \\
\midrule
Ours & \textcolor{red}{.0223} & \textcolor{red}{.8901} & \textcolor{red}{.8808} & \textcolor{red}{.9070} & \textcolor{red}{.9615} & \textcolor{red}{.8912} \\
\bottomrule
\end{tabular}}
\end{table}

%% file: tables/scope_ablation.tex
\begin{table}[!t]
\centering
\caption{Ablation study of Local and Global Mamba on ORSI-4199 attributes. The best result in each row is highlighted in \textcolor{red}{red}.}
\label{tab:scope-ablation}
\footnotesize
\setlength{\tabcolsep}{3.2pt}
\begin{tabular}{llccc}
\toprule
Attribute & Metric & Ours & All-local & All-global \\
\midrule
\multirow{3}{*}{BSO} & $S_m\uparrow$ & \textcolor{red}{.9083} & .8920 & .8927 \\
 & $F_\beta^w\uparrow$ & \textcolor{red}{.9218} & .9026 & .9033 \\
 & MS-SSIM$\uparrow$ & \textcolor{red}{.8593} & .8494 & .8503 \\
\addlinespace[1pt]
\multirow{3}{*}{SSO} & $S_m\uparrow$ & .8520 & \textcolor{red}{.8521} & .8514 \\
 & $F_\beta^w\uparrow$ & \textcolor{red}{.8222} & .8177 & .8144 \\
 & MS-SSIM$\uparrow$ & .9583 & \textcolor{red}{.9585} & .9580 \\
\addlinespace[1pt]
\multirow{3}{*}{NSO} & $S_m\uparrow$ & \textcolor{red}{.9100} & .9050 & .9065 \\
 & $F_\beta^w\uparrow$ & \textcolor{red}{.9118} & .9034 & .9063 \\
 & MS-SSIM$\uparrow$ & \textcolor{red}{.9573} & .9542 & .9551 \\
\addlinespace[1pt]
\multirow{3}{*}{ISO} & $S_m\uparrow$ & \textcolor{red}{.8975} & .8884 & .8898 \\
 & $F_\beta^w\uparrow$ & \textcolor{red}{.9169} & .9074 & .9093 \\
 & MS-SSIM$\uparrow$ & \textcolor{red}{.8720} & .8586 & .8603 \\
\addlinespace[1pt]
\multirow{3}{*}{CSO} & $S_m\uparrow$ & \textcolor{red}{.8917} & .8875 & .8886 \\
 & $F_\beta^w\uparrow$ & \textcolor{red}{.8954} & .8936 & .8951 \\
 & MS-SSIM$\uparrow$ & \textcolor{red}{.8567} & .8439 & .8457 \\
\bottomrule
\end{tabular}
\end{table}